\documentclass[10pt,twocolumn,letterpaper]{article}

\usepackage{cvpr}      %

\usepackage[T1]{fontenc} %
\usepackage{cuted} %

\usepackage{currfile} %

\usepackage{caption} %
\renewcommand{\paragraph}[1]{\vspace{.3em}\noindent\textbf{#1}}

\usepackage{microtype}

\usepackage{tikz}
\usetikzlibrary{positioning,arrows.meta,calc}
\usepackage{pgfplots}
\pgfplotsset{compat=1.17}
\usepackage{multirow}
\usepackage{colortbl}
\newcommand{\ours}{\textsc{SameQuota}\xspace}          %
\newcommand{\FP}{\textsc{FailParent}\xspace}           %
\newcommand{\PP}{\textsc{PassParent}\xspace}           %
\newcommand{\SR}{\textsc{SubtypeRate}\xspace}          %
\newcommand{\RD}{\textsc{Random}\xspace}               %
\newcommand{\CQ}{\textsc{CategoryQuota}\xspace}        %
\newcommand{\VS}{\textsc{ValueSwap}\xspace}            %
\newcommand{\FO}{\textsc{FailedOnly}\xspace}           %
\newcommand{\LR}{\textsc{LLMRewrite}\xspace}           %
\newcommand{\HO}{\textsc{HardObject}\xspace}           %
\newcommand{\conf}{Confirm-1400\xspace}                %
\newcommand{\contact}{Contact-200\xspace}              %
\newcommand{\idset}{ID-430\xspace}                     %

\definecolor{oiBlue}{HTML}{0072B2}
\definecolor{oiOrange}{HTML}{D55E00}
\definecolor{oiGreen}{HTML}{009E73}
\definecolor{findingbg}{HTML}{F3F5F8}
\definecolor{findingbar}{HTML}{0072B2}
\newcommand{\findingbox}[2]{%
  \par\smallskip\noindent
  \colorbox{findingbg}{%
    \parbox{\dimexpr\linewidth-2\fboxsep\relax}{%
      \vspace{1.5pt}\footnotesize\textbf{#1.}\enspace #2\vspace{1.5pt}%
    }%
  }\par\smallskip
}

\definecolor{cvprblue}{rgb}{0.21,0.49,0.74}
\usepackage[pagebackref,breaklinks,colorlinks,allcolors=cvprblue]{hyperref}

\def\confName{CVPR}
\def\confYear{2027}

\title{What to Practice and What to Trust: An Empirical Study of\\Data Choices in Text-to-Image Distillation}

\author{Zhao Wang\textsuperscript{*}, Chenghao Liu\textsuperscript{*\dag}, Junxiang Zeng\textsuperscript{*}, Junnan Nie, Yiming Du,\\
Fangxing Zheng, Jiachen Zhang, Kun Xu\textsuperscript{\ddag}, Songfang Huang\textsuperscript{\ddag}\\
Peking University}

\begin{document}
\maketitle
\begingroup
\makeatletter
\renewcommand{\@makefntext}[1]{\noindent#1}
\footnotetext{\scriptsize
  \begin{tabular}[t]{@{}l@{\kern.08em}l@{}}
    \textsuperscript{*} & Equal contribution.\quad\textsuperscript{\dag}Project leader.\\
    \textsuperscript{\ddag} & Corresponding authors: \texttt{syxu828@gmail.com}, \texttt{hsf@pku.edu.cn}.
  \end{tabular}}
\makeatother
\endgroup
\begin{strip}
\centering
\definecolor{inkSlate}{HTML}{1E293B}
\definecolor{mutedSlate}{HTML}{64748B}
\definecolor{tintBlue}{HTML}{EEF5FA}
\definecolor{tintWarm}{HTML}{FDF4EC}
\definecolor{tintSage}{HTML}{EEF7F2}
\begin{tikzpicture}[x=0.985cm, y=1cm, font=\scriptsize, text=inkSlate]

\fill[tintBlue, rounded corners=4pt] (0,0) rectangle (5.80,4.15);
\node[fill=oiBlue, text=white, font=\scriptsize\bfseries, rounded corners=2pt,
  inner xsep=4.5pt, inner ysep=1.8pt, anchor=north west] at (0.14,4.02)
  {Finding 1 $\cdot$ What to Trust: Larger Lever};
\node[font=\tiny, text=mutedSlate, anchor=north west] at (0.16,3.52)
  {Same 1,000 prompts: verified \emph{vs.} 1st teacher draw};

\begin{axis}[
  at={(0.68cm,0.50cm)}, anchor=south west,
  scale only axis, width=2.12cm, height=2.22cm,
  ybar, bar width=6.2pt, ymin=0, ymax=106, ytick={0,50,100},
  symbolic x coords={base,tchr,u1,ver}, xtick={base,tchr,u1,ver},
  xticklabels={Base, Tchr, 1st, \textbf{Ver.}},
  xticklabel style={font=\tiny, text=inkSlate},
  yticklabel style={font=\tiny, text=mutedSlate},
  ylabel={\conf (\%)}, ylabel style={font=\tiny, text=mutedSlate, yshift=-6pt},
  axis x line*=bottom, axis y line*=left,
  axis line style={draw=inkSlate!40}, tick style={draw=inkSlate!40},
  enlarge x limits=0.20,
]
\addplot[fill=inkSlate!30, draw=none, bar shift=0pt] coordinates {
  (base,56.8) (tchr,85.4) (u1,0) (ver,0)
};
\addplot[fill=oiOrange!85, draw=none, bar shift=0pt] coordinates {
  (base,0) (tchr,0) (u1,84.8) (ver,0)
};
\addplot[fill=oiBlue, draw=none, bar shift=0pt] coordinates {
  (base,0) (tchr,0) (u1,0) (ver,90.5)
};
\node[font=\tiny, text=inkSlate, anchor=south] at (axis cs:base,56.8) {56.8};
\node[font=\tiny, text=inkSlate, anchor=south] at (axis cs:tchr,85.4) {85.4};
\node[font=\tiny, text=oiOrange!90!black, anchor=south] at (axis cs:u1,84.8) {84.8};
\node[font=\tiny\bfseries, text=oiBlue, anchor=south] at (axis cs:ver,90.5) {90.5};
\end{axis}

\begin{axis}[
  at={(4.06cm,0.50cm)}, anchor=south west,
  scale only axis, width=1.22cm, height=2.22cm,
  xbar, bar width=6.2pt, xmin=0, xmax=6.5, xtick={0,2,4},
  symbolic y coords={cb,ge2,ge}, ytick={cb,ge2,ge},
  yticklabels={{CompB.},{GenEval\,2},{GenEval\textsubscript{P+C}}},
  yticklabel style={font=\tiny, text=inkSlate},
  xticklabel style={font=\tiny, text=mutedSlate},
  xlabel={Public $\Delta$ (pt)}, xlabel style={font=\tiny, text=mutedSlate, yshift=2pt},
  axis x line*=bottom, axis y line*=left,
  axis line style={draw=inkSlate!40}, tick style={draw=inkSlate!40},
  enlarge y limits=0.26,
  nodes near coords={+\pgfmathprintnumber[fixed,fixed zerofill,precision=1]{\pgfplotspointmeta}},
  nodes near coords style={font=\tiny\bfseries, text=oiBlue, anchor=west},
]
\addplot[fill=oiBlue, draw=none] coordinates {(4.15,ge) (1.04,ge2) (0.80,cb)};
\end{axis}

\fill[tintWarm, rounded corners=4pt] (6.02,0) rectangle (11.78,4.15);
\node[fill=oiOrange, text=white, font=\scriptsize\bfseries, rounded corners=2pt,
  inner xsep=4.5pt, inner ysep=1.8pt, anchor=north west] at (6.16,4.02)
  {Finding 2 $\cdot$ What to Practice: Mostly Tail};
\node[font=\tiny, text=mutedSlate, anchor=north west] at (6.18,3.52)
  {1,000 verified prompts per pool: tail \emph{vs.} CompBench};

\begin{axis}[
  at={(7.52cm,0.50cm)}, anchor=south west,
  scale only axis, width=3.85cm, height=2.05cm,
  xbar=1.0pt, bar width=3.8pt, area legend, xmin=0, xmax=100, xtick={0,50,100},
  symbolic y coords={cb,main,tail}, ytick={cb,main,tail},
  yticklabels={{CompB.\ avg.},{\conf},{5 objects}},
  yticklabel style={align=right, font=\tiny, text=inkSlate},
  xticklabel style={font=\tiny, text=mutedSlate},
  xlabel={Student score (\%)}, xlabel style={font=\tiny, text=mutedSlate, yshift=2pt},
  axis x line*=bottom, axis y line*=left,
  axis line style={draw=inkSlate!40}, tick style={draw=inkSlate!40},
  enlarge y limits=0.26,
  legend style={at={(0.48,1.03)}, anchor=south, legend columns=3, draw=none, fill=none, font=\tiny, column sep=3pt},
  legend image code/.code={\draw[#1, draw=none] (0cm,-0.06cm) rectangle (0.20cm,0.08cm);},
]
\addplot[fill=inkSlate!28, draw=none] coordinates {(59.6,tail) (86.9,main) (59.8,cb)};
\addlegendentry{Coarse}
\addplot[fill=inkSlate!55, draw=none] coordinates {(69.0,tail) (89.5,main) (59.7,cb)};
\addlegendentry{Random}
\addplot[fill=oiOrange!90!black, draw=none] coordinates {(73.5,tail) (90.5,main) (59.6,cb)};
\addlegendentry{\textbf{Fine}}
\node[font=\tiny\bfseries, text=oiOrange!90!black, anchor=west] at (axis cs:74.5,tail) {+14.0\,pt};
\node[font=\tiny, text=mutedSlate, anchor=west] at (axis cs:61.2,cb) {converge};
\end{axis}

\fill[tintSage, rounded corners=4pt] (12.00,0) rectangle (17.50,4.15);
\node[fill=oiGreen!85!black, text=white, font=\scriptsize\bfseries, rounded corners=2pt,
  inner xsep=4.5pt, inner ysep=1.8pt, anchor=north west] at (12.14,4.02)
  {Finding 3 $\cdot$ Verifier Blind Spots};
\node[font=\tiny, text=mutedSlate, anchor=north west] at (12.16,3.52)
  {Target for ``bottle on top of microwave'' (\FP pool)};

\node[inner sep=0pt] (imgBox) at (13.34,2.14)
  {\includegraphics[width=0.095\linewidth]{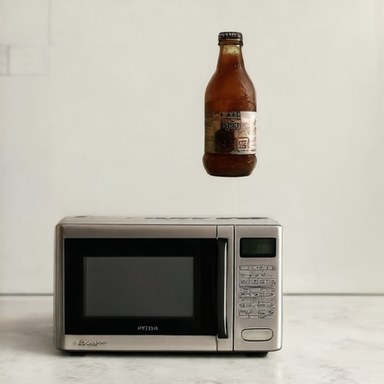}};
\draw[oiOrange, line width=0.8pt, dashed] ($(imgBox.south west)+(0.82,0.92)$) rectangle ($(imgBox.south west)+(1.14,1.54)$);
\draw[oiBlue, line width=0.8pt, dashed]   ($(imgBox.south west)+(0.22,0.12)$) rectangle ($(imgBox.south west)+(1.24,0.70)$);
\node[circle, fill=oiOrange, text=white, inner sep=0.55pt, font=\tiny\bfseries]
  at ($(imgBox.north east)+(-0.13,-0.13)$) {$\times$};
\node[font=\tiny\bfseries, text=oiOrange!95!black] at (13.34,1.08) {Box Target};
\node[font=\tiny, text=mutedSlate] at (13.34,0.80) {floats; student \textbf{84.2}};

\draw[-{Stealth[length=4pt]}, thick, draw=oiGreen!80!black] (14.28,2.14) -- (15.06,2.14)
  node[midway, above=2pt, font=\tiny\bfseries, text=oiGreen!80!black] {split};

\node[inner sep=0pt] (imgSplit) at (16.02,2.14)
  {\includegraphics[width=0.095\linewidth]{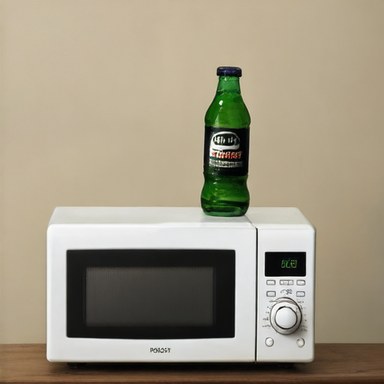}};
\draw[oiGreen!85!black, line width=1.0pt] (imgSplit.south west) rectangle (imgSplit.north east);
\node[circle, fill=oiGreen!85!black, text=white, inner sep=0.55pt, font=\tiny\bfseries]
  at ($(imgSplit.north east)+(-0.13,-0.13)$) {\checkmark};
\node[font=\tiny\bfseries, text=oiGreen!80!black] at (16.02,1.08) {Split Target};
\node[font=\tiny, text=mutedSlate] at (16.02,0.80) {touches; student \textbf{89.9}};

\node[fill=white, rounded corners=2pt, inner xsep=4pt, inner ysep=1.8pt, font=\tiny, text=inkSlate]
  at (14.75,0.32) {Student on \contact\ and GenEval\,2 ``on top of'': {\color{oiGreen!80!black}\textbf{+5.7}} and {\color{oiGreen!80!black}\textbf{+4.1}}};

\end{tikzpicture}
\captionsetup{hypcap=false}
\captionof{figure}{\textbf{Three findings on data choices in text-to-image distillation} (Flow-GRPO
teacher, SD3.5-Medium student, 1,000 prompts per pool). \textbf{Left:} on identical prompts, verified
targets ({\color{oiBlue}\textbf{blue}}) lift the student above the teacher's single draw on \conf, unlike
first draws ({\color{oiOrange}\textbf{orange}}), and raise all three public benchmark means (P+C:
Position+Counting). \textbf{Center:} fine-grained targeting (Fine, \ours) beats category quotas (Coarse,
\CQ) on ``5 objects'' and stays close to class-matched random prompts (Random, \RD); all three converge
on T2I-CompBench. \textbf{Right:} a box checker accepts a floating ``on top of'' target, whereas a split
detector--VLM verifier ({\color{oiGreen!80!black}\textbf{green}}) keeps a touching one and raises
student contact accuracy.}
\captionsetup{hypcap=true}
\label{fig:teaser}%
\end{strip}
\begin{abstract}
Supervised fine-tuning (SFT) on the images of an RL-tuned teacher can transfer compositional control to
a text-to-image student. It rests on two data choices: which prompts to practice and which teacher
images to trust. Recent pipelines target student weaknesses and filter with automated
verifiers, but change both choices at once, together with objectives and budgets.
Under a fixed prompt budget and SFT recipe, we compare ten prompt pools, from coarse quotas to
fine-grained subtype targeting, and contrast verified targets with first teacher draws on identical
prompts. Our testbed distills Flow-GRPO into SD3.5-Medium, with replications on a Sana student, a
Qwen-Image teacher and a second VLM judge. On pre-registered held-out position and counting prompts,
what to trust matters more than what to practice. Verification adds 5.7 points over first draws, lifts
the student above the teacher's single draw and also raises GenEval, GenEval\,2 and T2I-CompBench means. Fine-grained targeting beats coarse quotas by more than 3 points,
mostly on tail subtypes, yet stays within about one point of class-matched random prompts. The
verifier's blind spots, however, reach the student. A box checker accepts the teacher's floating ``on
top of'' images, and student contact accuracy falls by up to 13.8 points as these false passes
increase; a split detector--VLM verifier recovers 5.7 points. Distillation pipelines should
therefore verify targets before refining prompts, and audit the verifier itself.
\end{abstract}
\section{Introduction}
\label{sec:intro}

Reinforcement learning (RL) against programmatic rewards has made text-to-image models far more reliable
at counting and placing objects: Flow-GRPO reports a GenEval score of 95\% for SD3.5-Medium, up from
63\%~\cite{liu2025flowgrpo}. Running RL for every new model is costly, and a cheaper route is supervised
fine-tuning (SFT) on generated images that pass an automated check, as DreamSync does with its own
VLM-filtered samples~\cite{sun2025dreamsync}. Applied to the images of an RL-tuned teacher, this recipe
distills the teacher's skills into a student. It rests on two data choices: \emph{what to practice}
(which prompts to assemble) and \emph{what to trust} (which teacher images to keep as targets).

A growing line of work refines the first choice. It allocates prompts by weak category or object
concept~\cite{polyopd2026,agenticapt2026,generateanyscene2026,chen2026weakness} or rewrites failed
examples~\cite{lee2024llm2llm,zhao2024evolvedirector,mind2026}, while generated images are filtered with
reward models, detectors or vision-language models (VLMs)~\cite{dong2023raft,sun2025dreamsync,
xie2025sana15}. Because each pipeline changes both choices together with its objective and budget, their
relative value in text-to-image distillation is unknown. Evidence on verification is mixed: SFT on
filtered data slightly trails the base model in BiDPO~\cite{bidpo2026} and moves two E2V students in
opposite directions~\cite{e2v2026}, without an unfiltered control. Concurrent work on chart generation credits most of its
gain to verified construction~\cite{rsi2026}, but there the targets are rendered from code, whereas a
distilled student learns from teacher images whose errors a verifier may miss. The verifier itself thus
raises a second question. Box rules can confuse physical contact (``on'') with vertical offset
(``above'')~\cite{spatialreward2026}, and a teacher trained by RL against such rules may learn this
shortcut; teacher errors propagate in distillation~\cite{flowopd2026}, so verified targets may pass them on.

We answer these questions with a controlled study in which one data choice varies at a time
(\cref{fig:teaser}). A 1,000-prompt budget and one flow-matching SFT recipe stay fixed while ten prompt
pools span coarse category quotas to fine-grained subtype targeting, and verified targets are compared
with first teacher draws on identical prompts. Our fine-grained reference pool, \ours, keeps the prompts
the student is confirmed to fail and adds object-renamed prompts in proportion to each failed subtype.
The primary testbed distills Flow-GRPO into SD3.5-Medium~\cite{esser2024sd3}, and we replicate key
comparisons with a Linear-DiT Sana student~\cite{xie2025sana}, a Qwen-Image teacher~\cite{qwenimage} and
InternVL3.5 as a second judge~\cite{wang2025internvl35}. Besides GenEval\,2~\cite{geneval2},
GenEval~\cite{ghosh2023geneval} and T2I-CompBench~\cite{huang2023t2icompbench}, we score a
pre-registered held-out suite (\conf and \contact) whose balanced subtypes expose effects that public
totals average away.

On the held-out position and counting prompts of \conf, the two choices are far from equal. Trained on
the teacher's first draws, the student only matches its teacher (84.8 \vs 85.4); verifying the same
targets adds 5.7 points and lifts it above the teacher, and verification also raises all three public
benchmark means (\cref{sec:claim1}). The gain replicates with Sana as the student and with Qwen-Image as the
teacher (\cref{sec:scope}). Prompt granularity matters mainly on the tail. Fine-grained targeting beats
category-gap and hard-object quotas by more than 3 points on \conf, most of it on five-object counts. It
stays within about one point of class-matched random prompts, and the pools converge on T2I-CompBench,
whose counts mostly stay at four or fewer (\cref{sec:claim2}).

Target verification becomes a double-edged sword when the teacher and verifier share a blind spot.
Flow-GRPO, which was trained against GenEval's box-based rule, often draws ``on top of'' objects floating
in mid-air, and the box checker accepts them as SFT targets. A $2\times2$ design that crosses image source
with verifier shows that the box checker costs accuracy only when it filters teacher images. A split
verifier, which pairs object detection with a VLM contact check, recovers 5.7 points on \contact and 4.1
points on GenEval\,2 ``on top of'' prompts. The damage grows with the dose: when we swap split-verified
targets for floating false passes, student contact accuracy falls monotonically in every seed, by 13.8
points when they make up only 9.8\% of all targets (\cref{sec:claim3}).

We propose no new objective. Our contribution is a controlled account of when distillation data help
and when they pass on the teacher's errors:
\begin{itemize}[leftmargin=*,itemsep=1pt,topsep=2pt]
\item a testbed that varies prompt selection and target selection one at a time under a fixed budget and
  SFT recipe (\cref{sec:method,sec:setup});
\item a comparison of ten prompt pools with matched verification controls, replicated with a second
  student, teacher and judge (\cref{sec:claim1,sec:claim2,sec:scope});
\item a dose-controlled demonstration that a blind spot shared by teacher and verifier reaches the
  student (\cref{sec:claim3}).
\end{itemize}
\section{Related Work}
\label{sec:related}

\paragraph{Fine-tuning on verified generated data.}
Filtering synthetic targets before supervised fine-tuning is a standard way to improve alignment. In
language modeling, rejection-sampling pipelines such as STaR, RFT, RAFT, ReST and ReST-EM retain
generations with verified answers or high rewards~\cite{zelikman2022star,yuan2023rft,dong2023raft,
gulcehre2023rest,singh2024restem}. Analogous pipelines in text-to-image generation weight samples by
human-feedback rewards~\cite{lee2023aligning} or filter candidates with VLM judges and object
detectors, as in DreamSync~\cite{sun2025dreamsync}, SANA\,1.5~\cite{xie2025sana15},
EvoGen~\cite{evogen2025} and Fine-T2I~\cite{finet2i2026}. Yet SFT on filtered data slightly trails the
base model in BiDPO~\cite{bidpo2026} and moves two E2V students in opposite directions~\cite{e2v2026},
neither with an unfiltered control. Concurrently, RSI separates verified construction from failure
targeting for chart generation in unified multimodal models and attributes most of the gain to
verification~\cite{rsi2026}. We study teacher images in text-to-image distillation instead, hold the
prompt pool fixed to measure the verification gain across students and teachers, and trace the errors
that survive the verifier.

\paragraph{Distillation and post-training of flow models.}
Whereas reinforcement learning optimizes rewards on student rollouts~\cite{black2024ddpo,
liu2025flowgrpo} and Diffusion-DPO learns from preference pairs~\cite{wallace2024diffusiondpo},
on-policy distillation~\cite{agarwal2024gkd,gu2024minillm}
transfers capabilities from a stronger teacher, as adapted to flow matching by
Flow-OPD~\cite{flowopd2026} and combined with teacher-image losses and gap-weighted prompts in
Poly-OPD~\cite{polyopd2026}. Concurrent studies extend flow-model distillation to multi-teacher
ensembles, few-step generators, query schedules and cross-scale students~\cite{stepxopd2026,
phaseopd2026,focusopd2026,gfdopd2026,flowmapopd2026,welmimage2026,fourstep2026}. While those works
redesign the training objective, we hold plain flow-matching SFT fixed so differences reflect data
choices alone.

\paragraph{Weakness-targeted prompts.}
Choosing which examples to train on, and in which order, dates back to curriculum
learning~\cite{bengio2009curriculum}.
Recent text-to-image pipelines target weaknesses at different granularities. At the level of categories
or concepts, Generate Any Scene oversamples object concepts with large teacher--student
gaps~\cite{generateanyscene2026}, Chen \etal mine weak visual categories~\cite{chen2026weakness}, SpatialCritic builds a capability-graph
curriculum~\cite{spatialcritic2026}, and
Poly-OPD and APT shift prompt quotas toward weak categories or diagnosed
failures~\cite{polyopd2026,agenticapt2026}. At the level of single prompts or rollouts, EvolveDirector
expands VLM-judged weak prompts~\cite{zhao2024evolvedirector}, IPR relabels mistaken
generations~\cite{ipr2024}, and RL samplers pick high-variance~\cite{cgpo2026} or
capability-matched~\cite{adagrpo2026} prompts. In language models, LLM2LLM
rewrites failed examples~\cite{lee2024llm2llm}, MIND re-embeds localized errors in new
problems~\cite{mind2026}, and SwS synthesizes weakness-focused problems~\cite{sws2025}, though training
solely on failures can hurt~\cite{correcting2026}. Because each method evaluates under
its own budget and teacher, their relative value remains unknown; we implement representative
strategies under a single 1,000-prompt budget (\cref{tab:arms}) and compare them against \ours, which
targets multi-sample-confirmed single constraints.

\paragraph{Verifiers and what they miss.}
Automated compositional evaluation relies either on detector box rules as in GenEval and
T2I-CompBench~\cite{ghosh2023geneval,huang2023t2icompbench,huang2025t2icompbenchpp} or on VQA judges as
in TIFA, DSG and GenEval\,2~\cite{hu2023tifa,cho2024dsg,geneval2}, and such checks now serve as RL
rewards~\cite{vvr2026,cast2026}. Every proxy check, however, has blind spots.
SpatialReward notes that geometric rules struggle to tell ``on'' from ``above''~\cite{spatialreward2026},
and Flow-OPD notes qualitatively that flawed teacher images propagate during
distillation~\cite{flowopd2026}. In language models, Tiapkin \etal identify teacher hacking, in which a
student that imitates an imperfect teacher more closely drifts further from the true
distribution~\cite{tiapkin2025teacher}. Rather than treating such artifacts as anecdotal, we
decouple image source from verifier in a $2\times2$ experiment, map the false-pass dose-response curve,
and measure how much split detector--VLM verification recovers.

Works first public on or after 2026-09-15 are concurrent with ours~\cite{stepxopd2026,
spatialcritic2026,agenticapt2026,correcting2026,welmimage2026,fourstep2026,rsi2026,vvr2026,
cast2026,phaseopd2026,focusopd2026,gfdopd2026,flowmapopd2026}.
\section{Controlled Experimental Design}
\label{sec:method}

\begin{figure*}[t]
\centering
\definecolor{inkSlate}{HTML}{1E293B}
\definecolor{mutedSlate}{HTML}{64748B}
\definecolor{spineGray}{HTML}{475569}
\definecolor{tintGray}{HTML}{F1F5F9}
\definecolor{tintWarm}{HTML}{FFF5EB}
\definecolor{tintCool}{HTML}{EDF6FA}
\begin{tikzpicture}[
  x=0.985cm, y=1cm, font=\scriptsize, text=inkSlate,
  spine/.style={-{Stealth[length=5.2pt,width=4.0pt]}, line width=1.25pt, draw=spineGray},
  pill/.style={fill=#1, text=white, font=\tiny\bfseries, rounded corners=2pt, inner xsep=4.2pt, inner ysep=1.6pt},
]

\fill[tintGray, rounded corners=2.5pt] (0,0.04) rectangle (17.50,0.56);
\node[font=\tiny, text=inkSlate!88] at (8.75,0.30)
  {\textbf{Held fixed:}\quad 1,000 prompts per pool\quad$\cdot$\quad plain flow-matching SFT (\cref{eq:sft}), 2,000 steps, batch 8\quad$\cdot$\quad primary sweep: SD3.5-M, 8 seeds\quad$\cdot$\quad replications (Sana-1.6B, Qwen-Image, InternVL3.5): 3 seeds};

\fill[tintGray, rounded corners=4.5pt] (0,0.74) rectangle (3.25,4.28);
\node[pill=spineGray, anchor=north west] at (0.14,4.14) {1\enspace Failure Probing};
\node[font=\tiny, text=mutedSlate, anchor=north west] at (0.14,3.68) {3+3 student draws + VLM gate};

\node[inner sep=0pt] (p0) at (0.93,2.44) {\includegraphics[width=0.068\linewidth]{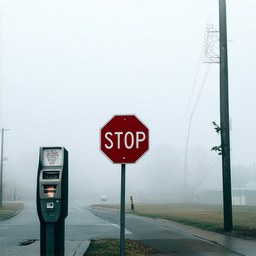}};
\node[circle, fill=oiOrange, text=white, inner sep=0.45pt, font=\tiny\bfseries]
  at ($(p0.north east)+(-0.10,-0.10)$) {$\times$};

\node[inner sep=0pt] (p3) at (2.32,2.44) {\includegraphics[width=0.068\linewidth]{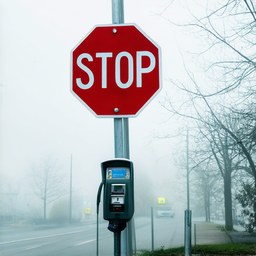}};
\node[circle, fill=oiOrange, text=white, inner sep=0.45pt, font=\tiny\bfseries]
  at ($(p3.north east)+(-0.10,-0.10)$) {$\times$};

\node[font=\tiny\itshape, text=inkSlate] at (1.625,1.50) {``stop sign \textbf{below} meter''};
\node[font=\tiny, text=oiOrange!90!black] at (1.625,1.10) {$\ge$4/6 fail $\rightarrow$ confirmed \texttt{below}};

\fill[tintWarm, rounded corners=4.5pt] (3.65,0.74) rectangle (8.55,4.28);
\node[pill=oiOrange, anchor=north west] at (3.80,4.14) {2\enspace Axis 1 $\cdot$ What to Practice};
\node[font=\tiny, text=mutedSlate, anchor=north west] at (3.80,3.68) {Prompt-pool granularity (10 pools, \cref{tab:arms})};

\fill[white, rounded corners=2.5pt] (3.81,2.68) rectangle (8.39,3.34);
\draw[draw=oiOrange!80, line width=0.75pt, rounded corners=2.5pt] (3.81,2.68) rectangle (8.39,3.34);
\node[font=\tiny\bfseries, text=oiOrange!90!black, anchor=west] at (3.92,3.12) {Fine Subtype (\ours\ + 2 pools)};
\node[font=\tiny, text=mutedSlate, anchor=west] at (3.92,2.85) {502 failed $+$ 498 renamed (\emph{``bench below frisbee''})};

\fill[white, rounded corners=2.5pt] (3.81,1.86) rectangle (8.39,2.52);
\node[font=\tiny\bfseries, text=inkSlate!88, anchor=west] at (3.92,2.30) {Coarse \& Uniform Quotas (4 pools)};
\node[font=\tiny, text=mutedSlate, anchor=west] at (3.92,2.03) {class gap, hard nouns, subtype rate, or random};

\fill[white, rounded corners=2.5pt] (3.81,1.04) rectangle (8.39,1.70);
\node[font=\tiny\bfseries, text=inkSlate!88, anchor=west] at (3.92,1.48) {Expansion Variants (3 pools)};
\node[font=\tiny, text=mutedSlate, anchor=west] at (3.92,1.21) {failed-only pool, swapped value, or LLM rewrite};

\fill[tintCool, rounded corners=4.5pt] (8.95,0.74) rectangle (13.95,4.28);
\node[pill=oiBlue, anchor=north west] at (9.09,4.14) {3\enspace Axis 2 $\cdot$ What to Trust};
\node[font=\tiny, text=mutedSlate, anchor=north west] at (9.09,3.68) {Teacher target filtering over 8--8--16 draws};

\node[inner sep=0pt] (tA) at (9.86,2.48) {\includegraphics[width=0.071\linewidth]{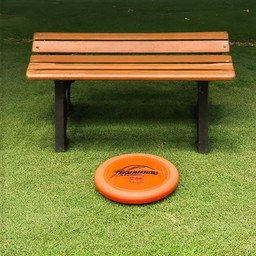}};
\node[circle, fill=oiOrange, text=white, inner sep=0.45pt, font=\tiny\bfseries]
  at ($(tA.north east)+(-0.10,-0.10)$) {$\times$};
\node[font=\tiny\bfseries, text=oiOrange!90!black] at (9.86,1.56) {First Draw};
\node[font=\tiny, text=mutedSlate, align=center] at (9.86,1.18) {unfiltered\\(inverted)};

\node[inner sep=0pt] (tB) at (11.45,2.48) {\includegraphics[width=0.071\linewidth]{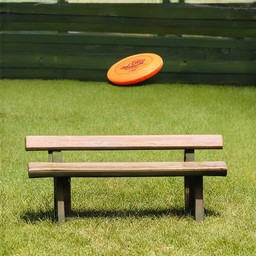}};
\draw[oiBlue, line width=1.15pt] (tB.south west) rectangle (tB.north east);
\node[circle, fill=oiGreen!85!black, text=white, inner sep=0.45pt, font=\tiny\bfseries]
  at ($(tB.north east)+(-0.10,-0.10)$) {\checkmark};
\node[font=\tiny\bfseries, text=oiBlue] at (11.45,1.56) {Box Verified};
\node[font=\tiny, text=mutedSlate, align=center] at (11.45,1.18) {first accepted $y$\\(contact blind)};

\node[inner sep=0pt] (tC) at (13.04,2.48) {\includegraphics[width=0.071\linewidth]{fig/img/contact/train-bottle-split.jpg}};
\draw[oiGreen!85!black, line width=1.05pt] (tC.south west) rectangle (tC.north east);
\node[circle, fill=oiGreen!85!black, text=white, inner sep=0.45pt, font=\tiny\bfseries]
  at ($(tC.north east)+(-0.10,-0.10)$) {\checkmark};
\node[font=\tiny\bfseries, text=oiGreen!80!black] at (13.04,1.56) {Split Verifier};
\node[font=\tiny, text=mutedSlate, align=center] at (13.04,1.18) {Detector + VLM\\(``on top of'' only)};

\fill[tintGray, rounded corners=4.5pt] (14.35,0.74) rectangle (17.50,4.28);
\node[pill=spineGray, anchor=north west] at (14.49,4.14) {4\enspace SFT and Evaluation};
\node[font=\tiny, text=mutedSlate, anchor=north west] at (14.49,3.68) {Same checkpoints, two tracks};

\fill[white, rounded corners=2.5pt] (14.51,2.54) rectangle (17.34,3.34);
\node[font=\tiny\bfseries, text=oiBlue, anchor=west] at (14.62,3.12) {Public Benchmarks};
\node[font=\tiny, text=mutedSlate, anchor=west] at (14.62,2.80) {GenEval\,2 $\cdot$ GenEval $\cdot$ CompB.};

\fill[white, rounded corners=2.5pt] (14.51,1.58) rectangle (17.34,2.38);
\node[font=\tiny\bfseries, text=inkSlate, anchor=west] at (14.62,2.16) {Held-out Diagnostic Suite};
\node[font=\tiny, text=mutedSlate, anchor=west] at (14.62,1.84) {\conf $\cdot$ \contact};

\node[font=\tiny, text=spineGray] at (15.925,1.15) {Qwen3-VL \& InternVL3.5};

\draw[spine] (3.27,2.51) -- (3.63,2.51);
\draw[spine] (8.57,2.51) -- (8.93,2.51);
\draw[spine] (13.97,2.51) -- (14.33,2.51);

\end{tikzpicture}
\caption{\textbf{Controlled design: one data choice varies at a time.} \textbf{(1)} Multi-sample probing
confirms student failures ({\color{oiOrange}$\times$}). \textbf{(2)} Ten 1,000-prompt pools range from
fine subtype preservation (\ours) to coarse quotas and rewrites (\cref{tab:arms}). \textbf{(3)} On
identical prompts, first draws ({\color{oiOrange}$\times$}) are compared with box-verified targets $y$
({\color{oiBlue}\textbf{blue}}); the split verifier ({\color{oiGreen!80!black}\textbf{green}}) is tested
on ``on top of'' prompts (\cref{sec:claim3}). \textbf{(4)} All students share one SFT recipe
(\cref{eq:sft}) and are scored on public and held-out suites.}
\label{fig:framework}
\end{figure*}

\begin{table*}[t]
\centering
\caption{\textbf{Ten prompt-pool constructions at 1,000 prompts} under one SFT recipe (8 seeds). Every
pool except \CQ shares the exact template-class counts (419 position, 380 count, 185 color, 16 object;
\CQ has 554, 405, 24, 17). %
New counts renamed or rewritten prompts and Failed counts failed originals. \ours anchors fine-grained
comparisons; \HO was added after reading the other nine.}
\label{tab:arms}
\small
\setlength{\tabcolsep}{4pt}
\begin{tabular}{@{}l l p{0.56\linewidth} rr@{}}
\toprule
Pool & In the spirit of & How the 1,000 prompts are chosen & New & Failed \\
\midrule
\multicolumn{5}{@{}l}{\emph{Fine-grained subtype preservation}} \\
\rowcolor{findingbg}
\ours & -- & Failed originals, plus $n_s$ random parents per failed subtype $s$, each renamed into up to 2
  new prompts & 498 & 502 \\ %
\FP & MIND~\cite{mind2026} & As \ours, but the parents are the failed originals themselves
  & 498 & 502 \\ %
\PP & -- & As \ours, but every parent is an original without a confirmed failure & 498 & 502 \\ %
\midrule
\multicolumn{5}{@{}l}{\emph{Uniform, rate-weighted and coarse quotas}} \\
\RD & -- & Random originals within \FP's template classes; new prompts as \ours & 498 & 41 \\ %
\SR & -- & Parents spread over subtypes by a softmax ($T=0.1$) of the
  student's per-subtype failure rates, random within a subtype & 498 & 60 \\ %
\CQ & \cite{polyopd2026,agenticapt2026} & Classes weighted by a softmax ($T=0.1$) of the
  teacher--student pass gap on 32 probe prompts per class, random within a class & 639 & 25 \\ %
\HO & \cite{generateanyscene2026} & As \RD, but every prompt contains one of the quarter of objects
  with the highest base error rate in its class & 498 & 44 \\ %
\midrule
\multicolumn{5}{@{}l}{\emph{Failed-only and expansion variants}} \\
\FO & \cite{correcting2026} & Failed originals only & 0 & 1,000 \\ %
\VS & -- & \FP, with the failed value of each new prompt swapped (\eg ``below'' $\to$ ``left of'')
  & 498 & 502 \\ %
\LR & \cite{lee2024llm2llm,zhao2024evolvedirector} & An LLM rewrites each of \FP's 250 parents (up to
  5 prompts per call, 2 calls); up to 4 per parent go to the teacher, and correct ones fill 2 per
  parent, topped up within the subtype & 498 & 502 \\ %
\bottomrule
\end{tabular}
\end{table*}

To isolate \emph{what to practice} from \emph{what to trust} without confounding objectives or budgets,
our testbed holds the training budget fixed at 1,000 prompts and varies one data axis at a time across
a four-stage pipeline (\cref{fig:framework}; Supp.\ Sec.\ S2).

\paragraph{Atomic constraints and programmatic checking.}
Diagnosing whole captions obscures whether a model fails on object identity, numeracy or spatial layout.
We represent each prompt $x$ by atomic constraints $c\in\mathcal{C}(x)$ specifying a kind (object,
count, color, position) and value, and call the kind--value pair of a position or count constraint its
\emph{subtype} (below, above, left of, right of; 2, 3, 4 or 5 objects). Following
GenEval~\cite{ghosh2023geneval}, a checker built on Mask2Former~\cite{cheng2022mask2former} and
CLIP~\cite{radford2021clip} labels each constraint as pass, fail or uncertain (counts match at confidence
$>0.9$; box centers satisfy the directional margin; scores within 0.05 of a threshold or inside the
margin are marked uncertain). Because it judges relations by box centers, we call it the \emph{box
checker}.

\paragraph{Noise-controlled failure probing.}
Because single-sample probing conflates genuine capability gaps with seed luck and detector noise, each
original prompt undergoes two-stage probing: 3 initial student draws, plus 3 retest draws whenever a
constraint fails at least twice. A constraint is a \emph{confirmed failure} only if it fails on $\ge 4$
of 6 images (at most 1 uncertain) and Gemini independently confirms the error on $\ge 2$ of the 3
retest images. Prompts with a confirmed failure are the \emph{failed originals}; those the teacher
cannot solve are dropped later, when targets are drawn.

\paragraph{Axis 1 (What to practice): spanning coarse-to-fine curation.}
We span the granularity spectrum of prior weakness-targeting methods with ten 1,000-prompt pools
(\cref{tab:arms}). Our fine-grained reference pool, \ours, retains 502 failed originals (123 position,
127 count, 131 color, 121 object failures). For each failed position or count subtype $s$ with $n_s$ failures, it
samples $n_s$ parents of that subtype uniformly and renames their objects by rule. This yields
245 new position and 253 new count prompts (\eg ``a dog below a table'' $\to$ ``a cup below a chair'').
Nine controlled alternatives hold the budget fixed while varying parent choice (\FP, \PP), subtype or class
quotas (\SR, \CQ, \HO), expansion (\FO, \VS, \LR), or uniform class-matched sampling (\RD).

\paragraph{Axis 2 (What to trust) and controlled SFT.}
To separate target quality from prompt solvability, verification controls train on the same 1,000
prompts with one of two targets: the teacher's first draw, or the first image the box checker accepts
within up to three rounds of 8, 8 and 16 draws. Position prompts that stay unsolved are redrawn with the
relation phrased in reverse, and the $<5\%$ still unsolved are dropped. Unless stated otherwise,
\emph{verified} means accepted by the box checker; \cref{sec:claim3} audits it. Every student then
minimizes plain flow-matching SFT loss~\cite{lipman2023flowmatching,
liu2023rectifiedflow,esser2024sd3},
\begin{equation}
  \mathcal{L}=\mathbb{E}_{(x,y),\,t,\,\epsilon}\big\|v_\theta(z_t,t,x)-(\epsilon-y)\big\|_2^2,
  \qquad z_t=(1-t)\,y+t\,\epsilon,
  \label{eq:sft}
\end{equation}
for $y$ the target latent, $\epsilon\sim\mathcal{N}(0,I)$ and logit-normal $t$, with no RL rollout or
auxiliary regularizer.
\section{Experimental Setup}
\label{sec:setup}

\paragraph{Models and training recipe.}
Our primary student is SD3.5-Medium~\cite{esser2024sd3} (attention LoRA~\cite{hu2022lora}, rank 32,
$\alpha=64$, Gaussian init, no dropout), complemented by heterogeneous Sana-1.6B-512px~\cite{xie2025sana}
(Linear-DiT, 32$\times$ DC-AE, Gemma-2-2B) trained on the exact same teacher image sets (\cref{sec:scope};
Supp.\ Sec.\ S5). The primary teacher is the public Flow-GRPO LoRA of SD3.5-Medium~\cite{liu2025flowgrpo}
(RL-trained on $\sim$50k GenEval-style prompts; 93.1 on GenEval in our evaluation), paired with Qwen-Image~\cite{qwenimage}
as a teacher from another model family. All runs share one pre-fixed recipe: AdamW~\cite{loshchilov2019adamw}
($\beta=(0.9,0.999)$, weight decay $10^{-4}$, clip 1.0), peak learning rate $3\times10^{-4}$ with 5\%
linear warm-up then linear decay to 0, batch 8, and 2,000 steps on 2 H100 GPUs ($\sim$13 min; raw step-2,000
weights evaluated as pre-registered, EMA tracked). Inference uses 512\,px and guidance 4.5 (40 steps on
SD3.5-Medium, 20 on Sana-1.6B).

\paragraph{Candidate prompts and pools.}
Training prompts come from one template family over 75 COCO object names ($\sim$2,150 position and
$\sim$1,650 count originals). Failure-based pools share one probe (21.0 GPU-hours; \CQ runs its own), and drawing and
checking teacher images takes 6.4 GPU-hours for \ours and 5.4 for \RD (Supp.\ Sec.\ S4). Every pool in \cref{tab:arms} keeps 1,000 verified
prompts and drops $<5\%$ of its candidate prompts; all except \CQ share identical template-class counts.
Each baseline follows the selection principle of the cited work, re-implemented under our budget and
checker. We pre-registered \FP as the primary pool, adopted \ours as the reference after reading the
first nine pools, and added \HO afterwards. \FP ties \ours (+0.2 on \conf), and every mean comparison on
\conf keeps its sign with \FP as the anchor.

\paragraph{Evaluation.}
\label{sec:protocol}
We evaluate the saved step-2,000 checkpoints on three unmodified public benchmarks.
GenEval\,2~\cite{geneval2} scores 800 prompts with one image each by the Soft-TIFA geometric mean (GM).
Classic GenEval~\cite{ghosh2023geneval} scores 553 prompts with 4 images each using its official
Mask2Former and CLIP checker, and we report its 6-task Overall and its Position+Counting mean.
T2I-CompBench~\cite{huang2023t2icompbench,huang2025t2icompbenchpp} scores 300 spatial and 300 numeracy
prompts with 4 images each using its official UniDet checker~\cite{zhou2022unidet}. We also report the 104 ``on top of'' atoms
of GenEval\,2 and \emph{practiced} subsets fixed before scoring. These keep the GenEval\,2 atoms on left
of, right of, under, on top of, two to five objects or a basic color, and the T2I-CompBench prompts with
a 2D relation or a largest count of two to five. On DrawBench~\cite{saharia2022imagen},
ImageReward~\cite{xu2023imagereward}, PickScore~\cite{kirstain2023pickapic} and diversity measure image
quality. Public totals mix practiced skills with unpracticed ones such as 3D relations and counts above
five, so we also score a pre-registered held-out suite, frozen on 2026-11-13 before any student was
scored (Supp.\ Sec.\ S3). \conf has 200 prompts for each of seven subtypes (below, above, left of, right
of, 3, 4 and 5 objects). They are spread over 8 cells that cross non-COCO objects, new phrasings and new
scenes, and they exclude 2,688 object pairs seen together in training; \contact has 200
physical-contact prompts. We also report \idset (430 prompts), a development
set used before the freeze. Qwen3-VL-8B~\cite{qwen3vl} scores 4 images per prompt on these suites via
Soft-TIFA~\cite{geneval2} with equal subtype and class weights; InternVL3.5-38B~\cite{wang2025internvl35}
re-scores 101,600 target questions, and Gemini audits a 240-prompt subset.

\paragraph{Coverage and seeds.}
All ten pools are scored on \conf, \contact and GenEval\,2 with 8 seeds, each drawn before the runs that
use it. T2I-CompBench covers seven pools; classic GenEval and the InternVL3.5 re-scoring cover \ours,
\RD, \HO and \CQ; Sana covers \ours, \RD and \CQ. First-draw controls exist for \ours and \FP on
SD3.5-Medium and for \ours with Sana and Qwen-Image. Classic GenEval, T2I-CompBench, InternVL3.5, Sana,
Qwen-Image, the contact factorial and the dose study use a fixed 3 of the 8 seeds. We report
every seed of every condition, with uncorrected paired $t$ intervals (Supp.\ Sec.\ S1).
\section{Results}
\label{sec:results}

\subsection{What to trust: target verification}
\label{sec:claim1}

\begin{table}[t]
\centering
\caption{\textbf{Verification raises \conf and all three public benchmark means} (SD3.5-M). Both
student rows train on the same 1,000 \ours prompts, using either the teacher's first draw (1st) or its
first checker-accepted image (ver.). $\Delta$: ver.\ minus 1st, with positive seeds in parentheses out of
8 (\conf, GE\,2) or 3 (GenEval, CompB.); bold: better student row. GE\,2: GenEval\,2; CompB.:
T2I-CompBench. Subtype breakdown in \cref{tab:claim2}; \contact in \cref{sec:claim3}.}
\label{tab:claim1}
\footnotesize
\setlength{\tabcolsep}{2.0pt}
\begin{tabular}{@{}l r r rr r@{}}
\toprule
& \conf & GE\,2 & \multicolumn{2}{c}{GenEval} & CompB. \\
\cmidrule(lr){4-5}
Model & Main $\uparrow$ & GM $\uparrow$ & All $\uparrow$ & P+C $\uparrow$ & Avg $\uparrow$ \\
\midrule
Base (SD3.5) & 56.81 & 18.19 & 68.27 & 43.78 & 43.82 \\ %
Teacher & 85.43 & 24.50 & 93.05 & 94.81 & 61.70 \\ %
\midrule
\ours, 1st & 84.79 & 22.25 & 88.02 & 80.28 & 58.83 \\ %
\rowcolor{findingbg}
\ours, ver. & \textbf{90.51} & \textbf{23.29} & \textbf{89.72} & \textbf{84.43} & \textbf{59.63} \\ %
\enspace $\Delta$ ($>0$) & +5.72(8) & +1.04(6) & +1.70(2) & +4.15(2) & +0.80(2) \\
\FP $\Delta$ & +4.56(8) & +1.90(8) & -- & -- & +0.82(2) \\ %
\bottomrule
\end{tabular}
\end{table}

\begin{table*}[t]
\centering
\caption{\textbf{Coarse quotas lose on the tail; fine-grained and random pools stay close.} Ten pools of
1,000 verified prompts on \conf, public GenEval\,2 and box-verified \contact (8 seeds; bold: best student
per column; split-verified contact control in \cref{sec:claim3}). $\Delta$ columns give seed-paired \ours
minus each pool, with positive seeds out of 8 and uncorrected 95\% $t$ intervals on \conf. \ours leads
\CQ and \HO by 3.0 to 3.6 points on \conf, whereas its leads over \RD (+1.1 on \conf, +0.7 on
GenEval\,2) have intervals that cross zero.}
\label{tab:claim2}
\footnotesize
\setlength{\tabcolsep}{1.65pt}
\begin{tabular}{@{}l rrr rrrr rrr r r l rr r@{}}
\toprule
& \multicolumn{3}{c}{\conf} & \multicolumn{4}{c}{Position} & \multicolumn{3}{c}{Count} & \multicolumn{3}{c}{\ours $-$ pool (\conf)} & \multicolumn{2}{c}{GenEval\,2} & Contact \\
\cmidrule(lr){2-4}\cmidrule(lr){5-8}\cmidrule(lr){9-11}\cmidrule(lr){12-14}\cmidrule(lr){15-16}\cmidrule(l){17-17}
Pool & Main $\uparrow$ & Pos. $\uparrow$ & Cnt. $\uparrow$ & below & above & left & right & 3 & 4 & 5 & $\Delta$ & $>0$ & $t$ interval & GM $\uparrow$ & $\Delta$ & Cont. $\uparrow$ \\
\midrule
Base & 56.81 & 68.33 & 45.30 & 57.25 & 73.68 & 77.02 & 65.38 & 66.40 & 45.61 & 23.88 & & & & 18.19 & & 94.00 \\ %
Teacher & 85.43 & 90.33 & 80.53 & 77.50 & 96.76 & 96.87 & 90.17 & 94.58 & 91.23 & 55.77 & & & & 24.50 & & 77.33 \\
\midrule
\rowcolor{findingbg}
\ours & \textbf{90.51} & 92.82 & \textbf{88.20} & 87.31 & 96.54 & \textbf{96.03} & 91.41 & 97.21 & 93.86 & \textbf{73.53} & & & & 23.29 & & 85.96 \\
\FP & 90.30 & 92.45 & 88.15 & 85.90 & 96.27 & 95.74 & 91.89 & 97.19 & 94.05 & 73.20 & +0.21 & 3/8 & [$-$0.94, +1.36] & 23.75 & $-$0.45 & \textbf{88.61} \\
\PP & 90.28 & 92.40 & 88.15 & 85.99 & \textbf{96.82} & 95.05 & 91.75 & 97.99 & 93.81 & 72.66 & +0.23 & 3/8 & [$-$0.78, +1.25] & 23.74 & $-$0.45 & 88.33 \\
\midrule
\SR & 89.89 & 91.91 & 87.87 & 84.36 & 95.50 & 95.57 & 92.22 & 97.40 & 93.09 & 73.12 & +0.62 & 5/8 & [$-$0.37, +1.61] & 22.26 & +1.03 & 86.94 \\
\FO & 89.69 & \textbf{92.87} & 86.51 & \textbf{87.97} & 96.31 & 95.17 & 92.03 & 95.97 & 92.31 & 71.26 & +0.82 & 7/8 & [+0.18, +1.45] & \textbf{24.14} & $-$0.84 & 86.41 \\
\RD & 89.46 & 91.21 & 87.70 & 82.61 & 96.07 & 93.89 & 92.28 & \textbf{98.36} & \textbf{95.74} & 69.00 & +1.05 & 6/8 & [$-$0.11, +2.22] & 22.57 & +0.72 & 86.19 \\
\LR & 89.25 & 92.15 & 86.35 & 84.32 & 96.23 & 95.68 & \textbf{92.36} & 97.65 & 92.25 & 69.16 & +1.26 & 7/8 & [+0.16, +2.36] & 23.79 & $-$0.50 & 87.55 \\
\VS & 88.54 & 91.84 & 85.25 & 84.05 & 96.53 & 94.95 & 91.81 & 97.85 & 92.78 & 65.13 & +1.97 & 7/8 & [+0.74, +3.20] & 23.32 & $-$0.03 & 86.32 \\
\HO & 87.47 & 91.94 & 83.00 & 83.36 & 96.17 & 95.88 & 92.34 & 97.16 & 92.81 & 59.02 & +3.04 & 8/8 & [+1.05, +5.04] & 22.59 & +0.70 & 86.38 \\
\CQ & 86.88 & 90.74 & 83.01 & 82.78 & 95.94 & 93.35 & 90.90 & 96.84 & 92.64 & 59.55 & +3.63 & 8/8 & [+1.81, +5.46] & 21.04 & +2.25 & 84.87 \\
\bottomrule
\end{tabular}
\end{table*}

On \conf, verified targets account for the student's entire margin over its same-base teacher
(\cref{tab:claim1}). SFT on the teacher's first draws already lifts SD3.5-Medium from 56.8 to 84.8, just
below the teacher's single draw (85.4). Verifying the same 1,000 targets adds 5.7 points in all 8 seeds
($t\,[+4.62,+6.82]$), and the pre-registered \FP pool shows a similar gain (+4.6).

The gain is largest where the teacher is weakest, consistent with verification acting mainly by
filtering out teacher errors. The teacher scores 80.5 on counts and 90.3 on positions, and verification
adds 8.0 and 3.4 points there. Counts gain more than positions in every replication, with
Qwen-Image as the teacher (+8.2 \vs +2.9), Sana as the student (+5.9 \vs +2.8) and InternVL3.5 as the
judge (+8.1 \vs +2.7).

Verification also raises all three public benchmark means. The largest public gain is on the practiced
Position+Counting tasks of classic GenEval (+4.2). On GenEval\,2, verification adds 1.0 GM for \ours and
1.9 for \FP (8/8 seeds) and brings the practiced atoms of \ours level with the teacher (56.2 \vs 56.1);
T2I-CompBench rises by 0.8. Students still trail the teacher on public totals. On GenEval\,2 the gap lies
in unpracticed atoms. On classic GenEval it lies entirely in Position (75.8 \vs 96.5), a task on which
Flow-GRPO was optimized with this benchmark's own detector rule~\cite{liu2025flowgrpo}.

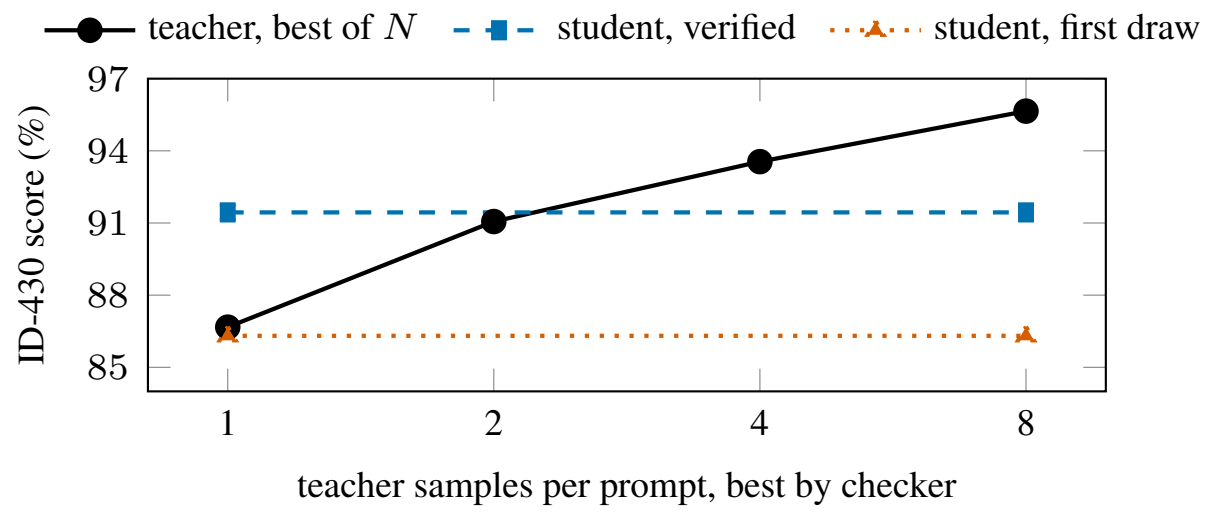
\begin{figure}[t]
\centering
\begin{tikzpicture}
\begin{axis}[
  width=0.95\linewidth, height=3.65cm,
  xmode=log, log basis x=2, xtick={1,2,4,8}, xticklabels={1,2,4,8},
  xlabel={teacher samples per prompt, best by checker}, ylabel={\idset score (\%)},
  ymin=84, ymax=97, ytick={85,88,91,94,97},
  legend columns=3, legend style={font=\scriptsize, at={(0.5,1.03)}, anchor=south, draw=none, fill=none,
    /tikz/every even column/.append style={column sep=4pt}},
  tick label style={font=\scriptsize}, label style={font=\scriptsize},
  every axis plot/.append style={thick},
]
\addplot[mark=*, color=black] coordinates {(1,86.67) (2,91.06) (4,93.55) (8,95.65)}; %
\addlegendentry{teacher, best of $N$}
\addplot[dashed, mark=square*, mark size=1.5pt, color=oiBlue] coordinates {(1,91.44) (8,91.44)}; %
\addlegendentry{student, verified}
\addplot[dotted, mark=triangle*, mark size=1.7pt, color=oiOrange] coordinates {(1,86.31) (8,86.31)}; %
\addlegendentry{student, first draw}
\end{axis}
\end{tikzpicture}
\caption{\textbf{One verified student draw matches the teacher's best of two.} On the development set
\idset, one draw of the verified \FP student ({\color{oiBlue}\textbf{blue}}) scores between the teacher's
checker-selected best of 2 and best of 4 (black), whereas first-draw SFT
({\color{oiOrange}\textbf{orange}}) stays at one teacher draw (base model: 58.6).}
\label{fig:bon}
\end{figure}

Verified distillation folds rejection sampling into a single student pass. On \idset, one draw of the
verified \FP student (91.4) falls between the teacher's checker-selected best of 2 (91.1) and best of 4
(93.6; \cref{fig:bon}). On a separate 200-prompt \conf subset judged by Gemini for one seed, the same
student also beats the teacher (+3.4 on target questions, +5.4 on whole prompts, paired). On DrawBench it
scores a higher ImageReward (1.22 \vs 1.06) at the cost of lower within-prompt diversity (79\% of the
teacher's; Supp.\ Sec.\ S5).

\findingbox{Finding 1 (Target verification)}{On identical prompts, verified targets add 5.7 points on
\conf and lift the student above the teacher's single draw. They also raise all three public benchmark means
(+0.8 to +1.7 on totals), and the gain holds with a second student and a second teacher.}

\subsection{What to practice: prompt granularity}
\label{sec:claim2}

Fine-grained pools beat coarse quotas on \conf, whereas the pools converge where evaluations rarely test
the tail (\cref{tab:claim2}). On \conf, \ours scores highest (90.5). Its 95\% $t$ intervals lie above zero
against category-gap quotas (\CQ, +3.6) and hard-object sampling (\HO, +3.0), and also against value
swapping, LLM rewrites and failed originals alone (+0.8 to +2.0). Its leads over class-matched \RD
(+1.1) and \SR (+0.6) have intervals that cross zero, and it ties \FP and \PP (+0.2). On public
benchmarks, the two quotas trail on different readouts. \ours beats \CQ by 2.25 GM on GenEval\,2 (7/8
seeds) but ties it on classic GenEval Position+Counting (+0.10). It beats \HO by 2.35 on
Position+Counting (3/3) but by only 0.70 GM on GenEval\,2. At budgets of 125, 250 and 500 prompts, the
margin of \ours over \RD is +2.3, +2.3 and $-$0.3 (Supp.\ Table\ S4), always below the 5.7-point
verification gain.

The gaps track how each pool covers the tail subtypes. \ours spends much of its count budget on the
subtype the student fails most often (``5 objects''), whereas \CQ samples count prompts at random within
the class and \HO chases hard object nouns. Both reach only about 59 on ``5 objects'' against 73.5 for
\ours, and they also trail on ``below'' (82.8 and 83.4 \vs 87.3). Class-level allocation further leaves
\CQ with only 24 color prompts (\vs 185 elsewhere), and the two color tasks (79.9 \vs 85.5) account for 1.9 of
its 2.0-point deficit in classic GenEval Overall. Where tail counts are absent, the pools
converge. The seven pools scored on T2I-CompBench, whose numeracy prompts mostly test at most four
objects (253/300), lie within 59.5--60.2. \ours, \RD and \CQ also lie within 84.3--84.5 on classic GenEval
Position+Counting, which tests 2 to 4 objects (Supp.\ Sec.\ S5).

\findingbox{Finding 2 (Prompt granularity)}{Fine-grained targeting beats category and hard-object quotas
by 3.0 to 3.6 points on \conf, with most of the gap on tail subtypes (``5 objects'', ``below''). It stays
within about one point of class-matched random prompts, and the pools converge on benchmarks dominated by
small counts.}

\subsection{Verifier blind spots reach the student}
\label{sec:claim3}

\begin{figure*}[t]
\centering
\newcommand{\cimgmark}[3]{%
  \begin{tikzpicture}[baseline=(img.south)]
    \node[inner sep=0pt] (img) {#1};
    \node[circle, fill=#2, text=white, inner sep=0.5pt, font=\tiny\bfseries]
      at ($(img.north east)+(-0.14,-0.14)$) {#3};
  \end{tikzpicture}%
}
\newcommand{\cimgno}[1]{\cimgmark{#1}{oiOrange!90!black}{$\times$}}
\newcommand{\cimgyes}[1]{\cimgmark{#1}{oiGreen!85!black}{\checkmark}}
\newcommand{\clab}[1]{{\scriptsize #1}}
\setlength{\tabcolsep}{1.2pt}
\renewcommand{\arraystretch}{0.55}
\begin{tabular}{@{}r cc@{\hspace{5pt}}cc@{\hspace{5pt}}cc@{}}
& \multicolumn{2}{c}{\clab{bottle on top of microwave}} & \multicolumn{2}{c}{\clab{fork on top of cake}}
  & \multicolumn{2}{c}{\clab{vase on top of chair}} \\[1pt]
\rotatebox{90}{\clab{\quad(a) training targets}} &
\cimgno{\includegraphics[width=0.130\linewidth]{fig/img/contact/train-bottle-box.jpg}} & \cimgyes{\includegraphics[width=0.130\linewidth]{fig/img/contact/train-bottle-split.jpg}} & \cimgno{\includegraphics[width=0.130\linewidth]{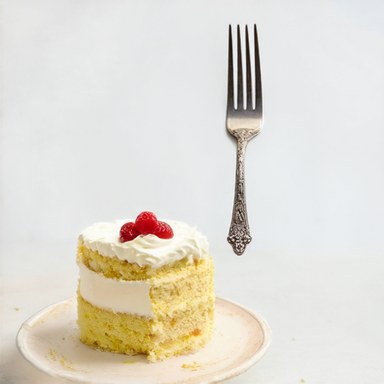}} & \cimgyes{\includegraphics[width=0.130\linewidth]{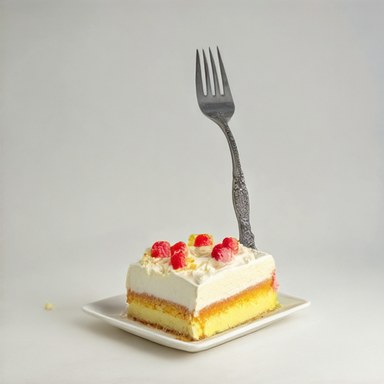}} &
\cimgno{\includegraphics[width=0.130\linewidth]{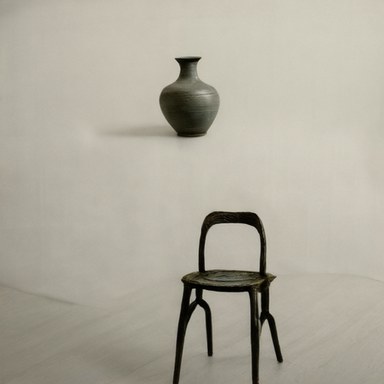}} & \cimgyes{\includegraphics[width=0.130\linewidth]{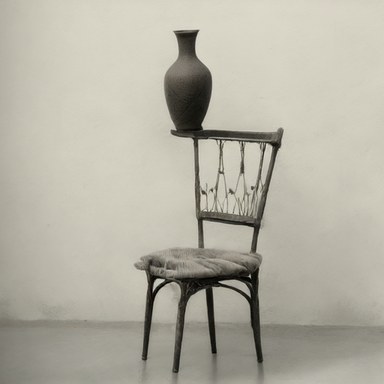}} \\[1pt]
& \clab{box checker} & \clab{split verifier} & \clab{box checker} & \clab{split verifier} & \clab{box checker}
  & \clab{split verifier} \\[5pt]
& \multicolumn{2}{c}{\clab{shoe on top of chair}} & \multicolumn{2}{c}{\clab{fork on top of toaster}}
  & \multicolumn{2}{c}{\clab{plant on top of stool \emph{(residual)}}} \\[1pt]
\rotatebox{90}{\clab{\quad\ (b) students' samples}} &
\cimgno{\includegraphics[width=0.130\linewidth]{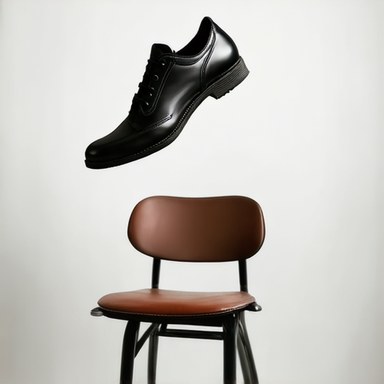}} & \cimgyes{\includegraphics[width=0.130\linewidth]{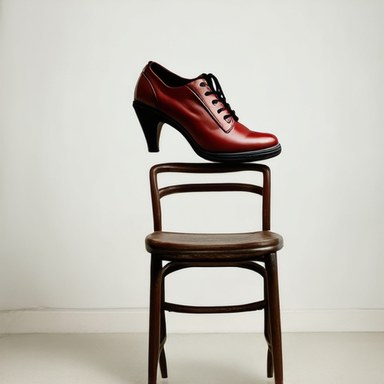}} & \cimgno{\includegraphics[width=0.130\linewidth]{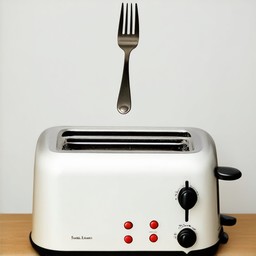}} & \cimgyes{\includegraphics[width=0.130\linewidth]{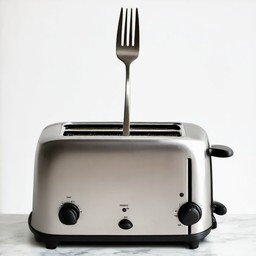}} &
\cimgno{\includegraphics[width=0.130\linewidth]{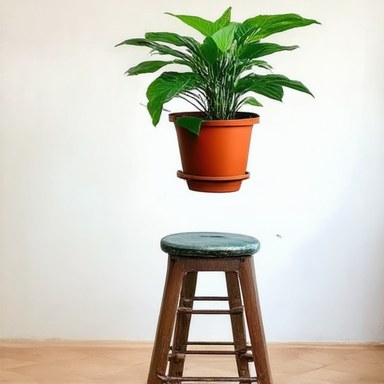}} & \cimgno{\includegraphics[width=0.130\linewidth]{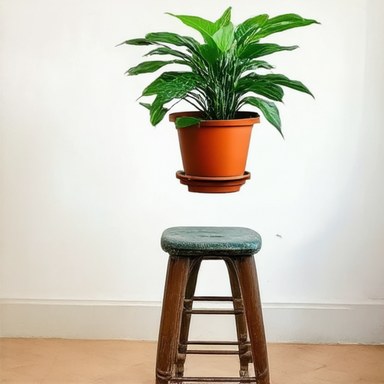}} \\[1pt]
& \clab{trained on box} & \clab{trained on split} & \clab{trained on box} & \clab{trained on split}
  & \clab{trained on box} & \clab{trained on split} \\
\end{tabular}
\caption{\textbf{Floating teacher targets reappear in students.} Badges mark floating objects
({\color{oiOrange!90!black}$\times$}) and visible contact ({\color{oiGreen!85!black}\checkmark}). (a) The
first three ``on top of'' training prompts by ID where the box checker accepts a floating teacher image
(left of each pair) and the split verifier keeps a touching one (right; \cref{tab:claim3}). (b) Student
samples on \contact (one seed): the first two prompts by ID where training on split-verified targets
removes the floating seen after box-checked training, and the first prompt where floating persists under
both (right).}
\label{fig:contact}
\end{figure*}

\begin{figure}[t]
\centering
\footnotesize
\setlength{\tabcolsep}{2.8pt}
\begin{tabular}{@{}l rr rr@{}}
\toprule
& \multicolumn{2}{c}{\contact} & \multicolumn{2}{c}{GenEval\,2 OnTop} \\
\cmidrule(lr){2-3}\cmidrule(l){4-5}
(a) Target source & Box $\uparrow$ & Split $\uparrow$ & Box $\uparrow$ & Split $\uparrow$ \\
\midrule
Teacher images & 84.18 & \textbf{89.90} & 25.45 & \textbf{29.50} \\ %
Base-model images & 95.30 & 95.15 & -- & -- \\
\midrule
Split $-$ box (teacher) &\multicolumn{2}{c}{\textbf{+5.72} (3/3)} & \multicolumn{2}{c}{\textbf{+4.05} (3/3)} \\
Interaction & \multicolumn{2}{c}{\textbf{+5.87} (3/3)} & \multicolumn{2}{c}{--} \\
\bottomrule
\end{tabular}
\par\vspace{3pt}
\begin{tikzpicture}
\begin{axis}[
  width=0.95\linewidth, height=3.35cm,
  xmin=-4, xmax=86, xtick={0,28,55,82}, xticklabels={0\%,28\%,55\%,82\%},
  xlabel={(b) Share of floating false-pass targets in 119 ``on top of'' prompts},
  ylabel={\contact (\%)},
  ymin=70, ymax=93, ytick={72,78,84,90},
  tick label style={font=\scriptsize}, label style={font=\scriptsize},
]
\addplot[only marks, mark=o, mark size=1.4pt, color=gray!80] coordinates {
  (0,91.22) (0,89.13) (0,89.34)
  (28,87.06) (28,82.35) (28,82.90)
  (55,86.66) (55,78.42) (55,80.34)
  (82,79.41) (82,77.63) (82,71.19)
};
\addplot[thick, mark=*, mark size=1.7pt, color=oiOrange] coordinates {
  (0,89.90) (28,84.10) (55,81.80) (82,76.07)
};
\node[font=\tiny\bfseries, text=oiOrange!90!black, anchor=south east] at (axis cs:80,74.2) {$-$13.8\,pt};
\end{axis}
\end{tikzpicture}
\caption{\textbf{The box checker passes teacher errors to the student.} (a) Student scores after
training on box-checked or split-verified ``on top of'' targets drawn from the teacher or the base model,
on \contact and the 104 ``on top of'' atoms of GenEval\,2 (3 seeds; Supp.\ Sec.\ S6). Interaction: the
split $-$ box gain with teacher images minus that with base-model images. The teacher itself scores 77.3
and 28.1. (b) Replacing split-verified targets with box-accepted floating images lowers \contact in
every seed (gray: seeds; {\color{oiOrange}\textbf{orange}}: mean, $-13.8$ at 82\%).}
\label{tab:claim3}
\end{figure}
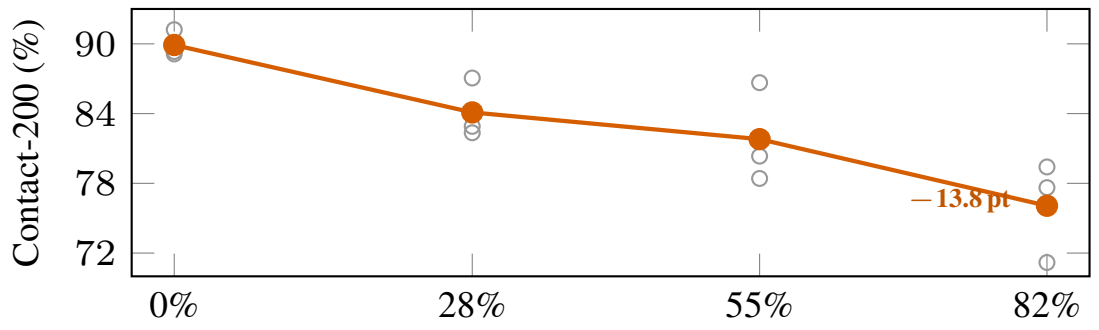

Because the box checker tests only whether box centers sit vertically apart, it accepts floating objects
as valid ``on top of'' targets. Flow-GRPO drops from the base model's 94.0 to 77.3 on \contact and from
45.3 to 28.1 on the 104 ``on top of'' atoms of GenEval\,2. Students distilled from its box-verified
targets fall to 84.9--88.6 on \contact across all ten pools (\cref{tab:claim2}) and to 23.0--27.4 on
GenEval\,2 ``on top of'' for the five pools reported there (Supp.\ Sec.\ S5). Verified \ours even scores
1.2 points below first-draw \ours on \contact (86.0 \vs 87.2).

A pre-registered $2\times2$ design that crosses image source with verifier shows that the box checker
costs accuracy only when it filters teacher images (\cref{tab:claim3}a). The design replaces 119
position prompts of the \FP pool with ``on top of'' prompts, and post-hoc Gemini auditing flags 40 of
the 119 box-accepted teacher targets as floating in mid-air (\cref{fig:contact}a). A split verifier, which
requires the detector to find both objects and Gemini to confirm contact (Supp.\ Sec.\ S6), lifts
students trained on teacher images from 84.2 to 89.9 on \contact (3/3 seeds, $t\,[+5.08,+6.35]$). It
also lifts them from 25.5 to 29.5 on GenEval\,2 ``on top of'' (3/3, $t\,[+1.83,+6.26]$), and
InternVL3.5 confirms the \contact gain (+5.2). With base-model images, the split verifier brings no gain
on \contact ($-$0.2), and the interaction there is positive in all 3 seeds (+5.87), so the gain is unlikely to reflect a preference of VLM judges for
VLM-selected images.

The more floating false passes the student sees, the lower its contact accuracy. Swapping 28\%, 55\% or
82\% of the 119 split-verified targets for box-accepted floating images lowers \contact from 89.9 to
84.1, 81.8 and 76.1, monotonically in every seed (\cref{tab:claim3}b). The 82\% level replaces only
9.8\% of all 1,000 targets, while \conf moves by only $-$1.0 to +0.5.

With a teacher that rarely floats objects, the box checker does no harm. Qwen-Image scores 99.0 on
\contact, and its students reach 96.7 with box-checked targets and 95.3 with first draws, consistent
with a blind spot that the teacher and verifier must share (compound ``below'' false passes in Supp.\
Sec.\ S7).

\findingbox{Finding 3 (Blind-spot inheritance)}{When teacher and verifier share a blind spot, the
verifier passes the teacher's errors to the student. Contact accuracy falls by 13.8 points when floating
false passes make up 9.8\% of all targets, and a split detector--VLM verifier reduces this transfer
(+5.7 on \contact, +4.1 on GenEval\,2 ``on top of'').}

\subsection{Replication across models and judges}
\label{sec:scope}

\begin{table}[t]
\centering
\caption{\textbf{Verification, coarse-quota and contact effects replicate on a second student and
under a second judge} (3 of the 8 seeds; positive seeds in parentheses; Supp.\ Sec.\ S5). \textbf{(a)}
Linear-DiT Sana-1.6B~\cite{xie2025sana} trained on the same teacher images; rows give \ours (ver.)
minus each control, and the last row uses the contact cells of \cref{tab:claim3}. \textbf{(b)} SD3.5-M
images re-scored by InternVL3.5-38B~\cite{wang2025internvl35} (95.9\% agreement with Qwen3-VL-8B).
OnTop: the 104 ``on top of'' atoms of GenEval\,2, reported as means only.}
\label{tab:generalize}
\footnotesize
\setlength{\tabcolsep}{1.05pt}
\begin{tabular}{@{}l rr rrr@{}}
\toprule
& \multicolumn{2}{c}{Held-out} & \multicolumn{3}{c}{Public} \\
\cmidrule(lr){2-3}\cmidrule(l){4-6}
(a) Sana-1.6B & Main $\uparrow$ & Cont. $\uparrow$ & GM $\uparrow$ & OnTop $\uparrow$ & CompB. $\uparrow$ \\
\midrule
Base & 58.42 & 89.88 & 13.71 & 36.90 & 45.06 \\ %
\rowcolor{findingbg}
\ours, ver. & \textbf{88.10} & 77.22 & \textbf{16.79} & 23.45 & \textbf{56.76} \\
\enspace $-$ first draw & +4.36(3) & +0.51(2) & +0.14(1) & $-$1.97 & +0.98(3) \\
\enspace $-$ \CQ & +3.46(3) & $-$0.93(1) & +0.76(2) & +0.29 & $-$0.48(0) \\
\enspace $-$ \RD & $-$0.22(1) & $-$1.29(0) & $-$0.10(1) & +0.83 & +0.06(2) \\
\midrule
Contact: split $-$ box & +0.00(1) & \textbf{+7.30}(3) & $-$0.05(1) & \textbf{+2.08}(2) & -- \\
\bottomrule
\end{tabular}
\par\vspace{3pt}
\setlength{\tabcolsep}{2.6pt}
\begin{tabular}{@{}l rr@{}}
\toprule
(b) InternVL3.5-38B (SD3.5-M) & Main $\uparrow$ & Cont. $\uparrow$ \\
\midrule
Ver.\ (89.5) $-$ 1st (84.1) & \textbf{+5.40} (3/3) & -- \\ %
\ours $-$ \CQ (84.4) & +5.12 (3/3) & -- \\
\ours $-$ \HO (85.1) & +4.35 (3/3) & -- \\
\ours $-$ \RD (87.7) & +1.74 (2/3) & -- \\
Split (91.5) $-$ box (86.3) & -- & \textbf{+5.22} (3/3) \\
\bottomrule
\end{tabular}
\end{table}

The verification gain, the coarse-quota gap and the contact repair all replicate on a second student
architecture (\cref{tab:generalize}a). Linear-DiT Sana-1.6B~\cite{xie2025sana} shares no backbone, VAE
or text encoder with Flow-GRPO, yet SFT on the same teacher images lifts it from 58.4 to 88.1 on \conf,
above the teacher's 85.4 and 4.4 points above first draws in 3/3 seeds. Verification also raises its
T2I-CompBench score by 1.0 in 3/3 seeds, whereas its GenEval\,2 gain is small (+0.14 GM). \ours beats
\CQ by 3.5 on \conf (3/3) and ties \RD ($-$0.2). Split verification recovers 7.3 points on \contact
(82.6 \vs 75.3, 3/3) and 2.1 on GenEval\,2 ``on top of''.

The gain also holds with a teacher from another model family. With Qwen-Image~\cite{qwenimage} as
the teacher, verification adds 5.5 points on \conf at 512\,px (83.9 \vs 78.4, 3/3 seeds) and 2.4 points when
1024\,px targets are downsampled to 512\,px (87.6 \vs 85.2, 3/3). Both students stay below Qwen-Image
itself (86.1 at 512\,px and 91.5 at 1024\,px, respectively).

A second judge keeps the sign of every margin (\cref{tab:generalize}b). On 101,600 SD3.5-M target
questions, InternVL3.5-38B~\cite{wang2025internvl35} agrees with Qwen3-VL-8B on 95.9\% of binary
answers ($r=0.892$). It credits verification with +5.4, \ours with +4.4 to +5.1 over coarse quotas and
split verification with +5.2, each in 3/3 seeds.
\section{Scope and Practical Guidelines}
\label{sec:limitations}

\paragraph{Scope and limitations.}
(1) \emph{Coverage:} all pools come from one COCO template family at 1,000 prompts. Students exceed the
50k-prompt Flow-GRPO teacher only on the practiced constraint families of \conf and \idset and trail it
on public totals (\cref{sec:claim1}). (2) \emph{Judges:} pairwise agreement with Qwen3-VL-8B is 95.9\% for InternVL3.5-38B
and 97--98\% for Gemini on a 240-prompt audit, and Mask2Former and UniDet provide non-VLM checks; still,
no human rated any image, and Gemini also takes part in probing and split verification.
(3) \emph{Post-hoc reference and hidden costs:} we pre-registered \FP and adopted \ours after reading the
first nine pools; pairwise $t$ intervals against \ours are uncorrected, and its leads over \RD cross zero
and vanish at 500 prompts and on T2I-CompBench. Equal training steps also hide unequal probing and
drawing cost, student diversity falls to 79\% of the teacher's, and the split verifier changes two things
at once (it drops the box rule and adds a VLM contact check). (4) \emph{Ranking:} on
GenEval\,2 GM and classic GenEval Overall, \ours leads \CQ by more (2.25 and 2.04) than verification adds
(1.04 and 1.70). (5) \emph{Blind spots:} we trace only ``on top of'' (plus a ``below'' audit), test the split verifier only on contact prompts, and run the $2\times2$ and dose studies with
3 seeds.

\paragraph{Practical guidelines for T2I distillation.}
(1) \emph{Verify targets before refining prompts:} on \conf,
filtering teacher draws adds 5.7 points, more than any prompt pool gains over class-matched random
sampling (at most 2.3). (2) \emph{Never let quotas starve tail subtypes:} class-level and hard-noun
quotas lose 3.0 to 3.6 points and most of it on ``5 objects'', whereas quotas at the failed-subtype level
and class-matched random sampling avoid this loss. (3) \emph{Audit the verifier on relations an RL
teacher may game:} a small Gemini audit exposed floating false passes, and a split verifier reduced
their transfer.

\section{Conclusion}
\label{sec:conclusion}

Varying one data choice at a time under a fixed budget and recipe shows that, on held-out position and
counting prompts, which teacher images to trust matters more than which prompts to practice. Verified
targets add 5.7 points on identical prompts and replicate with a second student, teacher and judge,
whereas fine-grained targeting helps mainly on tail subtypes. Verification has a risk of its own: a
verifier that shares the teacher's blind spot passes the teacher's errors to the student. A checker
that also shaped the teacher's RL reward should therefore be audited, on contact and on every other
relation that reward checks, before its verdicts become targets.

{
    \small
    \bibliographystyle{ieeenat_fullname}
    \bibliography{main}
}

\clearpage
\twocolumn[{\centering\Large\bfseries Appendix\par\vspace{1em}}]
\setcounter{section}{0}
\renewcommand{\theHsection}{appendix.\arabic{section}}
\setcounter{table}{0}
\renewcommand{\theHtable}{appendix.\arabic{table}}
\setcounter{figure}{0}
\renewcommand{\theHfigure}{appendix.\arabic{figure}}

\renewcommand{\thesection}{S\arabic{section}}
\renewcommand{\thetable}{S\arabic{table}}
\renewcommand{\thefigure}{S\arabic{figure}}

\section{Per-seed results}
\label{app:seeds}

\begin{table*}[t]
\centering
\caption{\textbf{Main score per seed} on \conf, every pool. Seeds s1--s8 in the order of
\cref{tab:seeds}. Base model 56.81 and teacher 85.43 (no training seed).}
\label{tab:seeds-abs}
\footnotesize
\setlength{\tabcolsep}{4pt}
\begin{tabular}{@{}l rrrrrrrr r@{}}
\toprule
Pool & s1 & s2 & s3 & s4 & s5 & s6 & s7 & s8 & Mean $\uparrow$ \\
\midrule
\ours & 89.22 & 90.49 & 91.16 & 90.82 & 90.85 & 89.97 & 90.75 & 90.82 & 90.51 \\ %
\FP & 90.77 & 91.17 & 90.16 & 87.98 & 89.74 & 90.60 & 90.87 & 91.10 & 90.30 \\
\PP & 90.24 & 88.57 & 90.66 & 91.78 & 90.99 & 90.54 & 90.75 & 88.70 & 90.28 \\
\SR & 89.95 & 90.61 & 92.12 & 89.47 & 90.44 & 87.82 & 89.95 & 88.77 & 89.89 \\
\FO & 89.90 & 89.29 & 89.71 & 90.76 & 89.80 & 88.78 & 89.99 & 89.32 & 89.69 \\
\RD & 88.98 & 89.55 & 87.51 & 90.56 & 91.33 & 88.12 & 88.62 & 90.98 & 89.46 \\
\LR & 90.79 & 88.26 & 90.40 & 88.51 & 88.24 & 88.76 & 89.61 & 89.44 & 89.25 \\
\VS & 88.72 & 87.87 & 87.13 & 89.25 & 88.11 & 87.76 & 91.41 & 88.10 & 88.54 \\
\HO & 88.41 & 86.24 & 83.44 & 88.72 & 88.98 & 88.41 & 85.87 & 89.67 & 87.47 \\
\CQ & 87.43 & 87.43 & 82.87 & 89.08 & 86.45 & 87.93 & 87.63 & 86.19 & 86.88 \\
\bottomrule
\end{tabular}
\end{table*}

\begin{table*}[t]
\centering
\caption{\textbf{\ours minus each pool, per seed} on \conf. Seeds in the order 24818887, 167930152,
249530483, 648258127, 659064548, 1315432225, 1844363854, 2111277986. The last two columns are 95\%
intervals: the $t$ interval over the 8 seed differences and the prompt bootstrap described in
\cref{app:seeds}. Post hoc: \ours became the reference pool after the pre-registered comparisons.}
\label{tab:seeds}
\footnotesize
\setlength{\tabcolsep}{3.5pt}
\begin{tabular}{@{}l rrrrrrrr r ll@{}}
\toprule
\ours $-$ & s1 & s2 & s3 & s4 & s5 & s6 & s7 & s8 & Mean & Seed $t$ interval & Prompt interval \\
\midrule
\CQ & +1.79 & +3.06 & +8.29 & +1.74 & +4.39 & +2.04 & +3.11 & +4.63 & +3.63 & [+1.81, +5.46] & [+3.19, +4.13] \\ %
\HO & +0.81 & +4.25 & +7.73 & +2.10 & +1.86 & +1.56 & +4.87 & +1.15 & +3.04 & [+1.05, +5.04] & [+2.56, +3.51] \\ %
\VS & +0.50 & +2.62 & +4.03 & +1.57 & +2.74 & +2.21 & $-$0.66 & +2.73 & +1.97 & [+0.74, +3.20] & [+1.54, +2.43] \\
\LR & $-$1.57 & +2.23 & +0.77 & +2.31 & +2.60 & +1.21 & +1.14 & +1.39 & +1.26 & [+0.16, +2.36] & [+0.89, +1.67] \\
\RD & +0.24 & +0.95 & +3.65 & +0.25 & $-$0.48 & +1.85 & +2.12 & $-$0.15 & +1.05 & [$-$0.11, +2.22] & [+0.60, +1.52] \\
\FO & $-$0.67 & +1.20 & +1.46 & +0.06 & +1.05 & +1.20 & +0.76 & +1.51 & +0.82 & [+0.18, +1.45] & [+0.38, +1.27] \\
\SR & $-$0.72 & $-$0.12 & $-$0.96 & +1.34 & +0.40 & +2.15 & +0.80 & +2.06 & +0.62 & [$-$0.37, +1.61] & [+0.22, +1.03] \\
\PP & $-$1.01 & +1.92 & +0.50 & $-$0.96 & $-$0.14 & $-$0.57 & $-$0.00 & +2.13 & +0.23 & [$-$0.78, +1.25] & [$-$0.20, +0.65] \\
\FP & $-$1.54 & $-$0.68 & +1.00 & +2.84 & +1.11 & $-$0.63 & $-$0.12 & $-$0.28 & +0.21 & [$-$0.94, +1.36] & [$-$0.15, +0.56] \\
\bottomrule
\end{tabular}
\end{table*}

\paragraph{Scoring.}
All 8 seeds of every pool are reported, with scores in \cref{tab:seeds-abs} and differences in
\cref{tab:seeds}; no scoring item was missing for any model.
A model's main score averages, per prompt, the target question's Soft-TIFA score over its 4 images,
then the prompts of each subtype, the subtypes equally within position and within count, and the two
classes equally.
\cref{fig:effects} plots the differences of \cref{tab:seeds} on one axis with the verification gains.

\begin{figure}[t]
\centering
\begin{tikzpicture}
\begin{axis}[
  width=0.62\linewidth, height=7.4cm,
  y dir=reverse, ymin=-0.7, ymax=16.6,
  xmin=-1.5, xmax=11, xtick={0,2,4,6,8,10},
  xlabel={paired difference on \conf (points)},
  ytick={0,1,...,16},
  yticklabels={
    \textbf{Verified $-$ first draw},
    {\ours, SD3.5-M}, {\FP, SD3.5-M}, Sana-1.6B student$^\dagger$, Qwen-Image teacher$^\dagger$,
    Qwen-Image 1024\,px$^\dagger$, InternVL3.5 judge$^\dagger$,
    \textbf{\ours $-$ pool},
    \CQ, \HO, \VS, \LR, \RD, \FO, \SR, \PP, \FP},
  yticklabel style={font=\scriptsize}, major y tick style={draw=none},
  tick label style={font=\scriptsize}, label style={font=\scriptsize},
  clip=false,
  error bars/error bar style={thick},
]
\draw[gray!70, dashed] (axis cs:0,-0.7) -- (axis cs:0,16.6);
\draw[gray!35] (axis cs:-1.5,6.55) -- (axis cs:11,6.55);
\addplot[only marks, mark=square*, mark size=1.6pt, color=oiBlue,
  error bars/.cd, x dir=both, x explicit] coordinates {
  (5.72,1) += (1.10,0) -= (1.10,0)
  (4.56,2) += (1.07,0) -= (1.08,0)
  (4.36,3) += (3.67,0) -= (3.68,0)
  (5.50,4) += (4.01,0) -= (4.00,0)
  (2.39,5) += (2.85,0) -= (2.85,0)
  (5.40,6) += (1.44,0) -= (1.45,0)
};
\addplot[only marks, mark=*, mark size=1.5pt, color=black,
  error bars/.cd, x dir=both, x explicit] coordinates {
  (3.63,8) += (1.83,0) -= (1.82,0)
  (3.04,9) += (2.00,0) -= (1.99,0)
  (1.97,10) += (1.23,0) -= (1.23,0)
  (1.26,11) += (1.10,0) -= (1.10,0)
  (1.05,12) += (1.17,0) -= (1.16,0)
  (0.82,13) += (0.63,0) -= (0.64,0)
  (0.62,14) += (0.99,0) -= (0.99,0)
  (0.23,15) += (1.02,0) -= (1.01,0)
  (0.21,16) += (1.15,0) -= (1.15,0)
};
\tikzset{val/.style={anchor=west, font=\tiny, inner sep=1.5pt}}
\node[val] at (axis cs:6.82,1) {+5.72};
\node[val] at (axis cs:5.63,2) {+4.56};
\node[val] at (axis cs:8.03,3) {+4.36};
\node[val] at (axis cs:9.51,4) {+5.50};
\node[val] at (axis cs:5.24,5) {+2.39};
\node[val] at (axis cs:6.84,6) {+5.40};
\node[val] at (axis cs:5.46,8) {+3.63};
\node[val] at (axis cs:5.04,9) {+3.04};
\node[val] at (axis cs:3.20,10) {+1.97};
\node[val] at (axis cs:2.36,11) {+1.26};
\node[val] at (axis cs:2.22,12) {+1.05};
\node[val] at (axis cs:1.45,13) {+0.82};
\node[val] at (axis cs:1.61,14) {+0.62};
\node[val] at (axis cs:1.25,15) {+0.23};
\node[val] at (axis cs:1.36,16) {+0.21};
\end{axis}
\end{tikzpicture}
\caption{\textbf{Verification moves \conf more than prompt selection.} Mean paired differences on \conf
with uncorrected 95\% $t$ intervals over seeds. Top: the same 1,000 \ours prompts with verified \vs first-draw
teacher targets, on the main testbed (8 seeds; \FP: its own prompts) and in the replications of Sec.~5.4 of
the main paper ($^\dagger$3 seeds; Qwen-Image 1024\,px: the 933 prompts with an accepted image, \cref{app:readouts}). Bottom: \ours minus each other pool (8 seeds, \cref{tab:seeds}; post hoc).
Both 8-seed verification gains exceed every pool difference; the 3-seed Qwen-Image 1024\,px gain is smaller
than the gaps to the two coarse quotas.}
\label{fig:effects}
\end{figure}

\paragraph{Intervals.}
The seed interval is the 95\% $t$ interval of the 8 seed differences (7 degrees of
freedom). The prompt interval redraws the prompts of each subtype with replacement (2,000 draws,
the same draws for both models and every seed), averages the difference over seeds in each draw and
takes the 2.5th and 97.5th percentiles; it covers prompt sampling only, not training noise.

\paragraph{Seeds and pre-registration.}
Every seed was drawn with \texttt{secrets.randbelow(2**31)} before the runs that use it: three on
2026-11-07 and five on 2026-11-13. %
The pre-registered comparisons on \conf were \FP against \SR, \PP, \LR and \ours; the comparisons of
\ours against the other pools were computed after we adopted it as the reference pool. \HO was added on
2026-11-15, and its criterion (\ours higher on the mean over the 8 seeds) was fixed before it was trained. %

\section{Probe, pools and teacher draws}
\label{app:arms}

\paragraph{Probe.}
A constraint is a confirmed failure when it fails on at least 2 of the first 3 student images, on at
least 4 of all 6, is uncertain on at most 1, and Gemini (gemini-3.8-flash) calls it drawn wrong on at
least 2 of the 3 retest images. %
The probe drew 57,354 student images over the pool, with 9,957 Gemini judgments. %

\paragraph{Assembling \ours.}
The quotas are \FP's failed originals per subtype: above 20, below 73, left of 5, right of 25; 3
objects 15, 4 objects 38, 5 objects 74 (no failure on 2 objects). For each subtype the eligible
originals are listed in a fixed hash order, and the first $\max(\lceil 1.5n\rceil, n+10)$ are kept
for a quota of $n$.
Each parent gets up to 3 renamed candidates, which replace its objects with distinct ones from the 75
object names of the templates (none of the parent's) and keep its other constraints and scene. A
candidate is dropped if its combination is ambiguous, if it repeats a held-out, test, benchmark or
confirmation-set prompt, a failed original or another prompt of the pool, or if it names together the
two objects of a test position prompt or of a two-object confirmation-set prompt. The teacher draws every
candidate, and parents are taken in order when at least 2 of their candidates have a verified image;
their first 2 verified candidates become the new prompts, except that the last parent of ``below''
and of ``5 objects'' gives 1 each, so that these 498 new prompts and the 502 failed originals of \FP
total 1,000. Of the 250 parents taken, 36 are failed originals themselves.

\paragraph{Pools.}
\SR gives position parents to above, below, left of and right of in the ratio 10/74/12/27, and count
parents to 2, 3, 4 and 5 objects in the ratio 3/8/27/89. %
\CQ uses a separate probe of 32 prompts per class with 3 images from each model, and allocates by
the teacher--student pass gap. %
\LR asks the LLM for up to 5 prompts per call and at most 2 calls per parent, and sends up to 4 valid
prompts per parent to the teacher: of 2,150 proposals, 1,691 were valid and 39\% kept
the failed constraint; 805 went to the teacher. %
For 26 parents too few rewrites were drawn correctly, so the 34 missing prompts were filled with spare
correct rewrites of other parents of the same subtype. %
\HO marks, in each class, the quarter of objects with the highest base error rate as hard (19 of
about 75); hard objects make up 61\% of object mentions in position prompts against 23\% in \RD.
Generate Any Scene ranks concepts by the teacher--student gap; as the teacher passes 93\% of GenEval,
\HO uses the base model's error rate as a proxy for that gap.
\cref{fig:pools} shows new prompts that \ours and \LR built from the same failed originals.

\begin{figure*}[t]
\centering
\scriptsize
\setlength{\tabcolsep}{4pt}
\begin{tabular}{@{}l p{0.72\linewidth} l@{}}
\toprule
Prompt & Text & Subtype \\
\midrule
\rowcolor{findingbg} Failed original & a photo of a boat and a dog, the boat is below the dog, in a snowy field. & below \\
\enspace \ours & a photo of a train and a skateboard, the train is below the skateboard, in a snowy field. & below \\
\enspace \ours & a photo of a sandwich and a scissors, the sandwich is below the scissors, in a snowy field. & below \\
\enspace \LR & a photo of a bicycle and a kite, the bicycle is below the kite, at sunset on an open hill. & below \\
\enspace \LR & a photo of a suitcase and a clock, the suitcase is below the clock, in a sunlit room with bare walls. & below \\
\midrule
\rowcolor{findingbg} Failed original & a photo of an umbrella and a scissors, the umbrella is below the scissors. & below \\
\enspace \ours & a photo of a toilet and a toaster, the toilet is below the toaster. & below \\
\enspace \ours & a photo of a pizza and a surfboard, the pizza is below the surfboard. & below \\
\enspace \LR & a photo of a tie and a suitcase, the tie is below the suitcase. & below \\
\enspace \LR & a photo of a laptop and a tv remote, the laptop is to the right of the tv remote. & right of$^\dagger$ \\
\midrule
\rowcolor{findingbg} Failed original & a photo of five sinks, on a wooden floor. & 5 objects \\
\enspace \ours & a photo of five oranges, on a wooden floor. & 5 objects \\
\enspace \ours & a photo of five tvs, on a wooden floor. & 5 objects \\
\enspace \LR & a photo of three couches, in a sunlit room with bare walls. & 3 objects$^\dagger$ \\
\enspace \LR & a photo of five benches, in front of a brick wall. & 5 objects \\
\midrule
\rowcolor{findingbg} Failed original & a photo of four sandwiches, on a plain white background. & 4 objects \\
\enspace \ours & a photo of four handbags, on a plain white background. & 4 objects \\
\enspace \ours & a photo of four baseball bats, on a plain white background. & 4 objects \\
\enspace \LR & a photo of four donuts, on a stone plaza. & 4 objects \\
\enspace \LR & a photo of three bananas, on a plain white background. & 3 objects$^\dagger$ \\
\bottomrule
\end{tabular}
\caption{\textbf{New prompts built from the same failed originals.} Of the 13 failed originals that are parents of new prompts in both \ours and \LR, the first two position and the first two count parents by ID (shaded), each with its new prompts in the two 1,000-prompt pools. \ours renames the objects and keeps the failed subtype and the scene; the LLM rewrites of \LR may change the scene and the subtype ($^\dagger$).}
\label{fig:pools}
\end{figure*}

\paragraph{Teacher draws.}
Of the 1,000 prompts of \ours, the teacher's first round of 8 images gave the verified image for 917,
the second round for 29, the third for 28, and the rephrased redraws for 26. %
Across the nine pools other than \HO, 91 of the 9,925 prompts sent to the teacher got no verified image. %

\section{Confirmation set}
\label{app:confirm}

\paragraph{\conf.}
Each subtype has 200 prompts in 8 cells of 25, crossing three changes: all objects from a list outside
the checker's COCO classes or from the training vocabulary; one of several new phrasings or the
training caption; a new scene or a training scene. The cell with no change uses the training
templates. A position or contact prompt never pairs two objects that appear together in a prompt of any
training pool or earlier test set (2,688 pairs excluded), and no prompt repeats one of those
prompts.
\cref{fig:confirmset} lists one prompt from each cell for two subtypes.

\begin{figure*}[t]
\centering
\scriptsize
\setlength{\tabcolsep}{4pt}
\begin{tabular}{@{}ccc p{0.38\linewidth} p{0.38\linewidth}@{}}
\toprule
New obj. & New phr. & New scene & ``below'' & ``5 objects'' \\
\midrule
-- & -- & -- & a photo of a fork and an oven, the fork is below the oven, in front of a brick wall. & a photo of five bananas, on a wooden floor. \\
-- & -- & \checkmark & a photo of a suitcase and a computer keyboard, the suitcase is below the computer keyboard, in a wheat field. & a photo of five buses, in a wheat field. \\
-- & \checkmark & -- & a toothbrush below a cake, on a wooden floor. & an image showing five motorcycles, at sunset on an open hill. \\
-- & \checkmark & \checkmark & a picture where the vase is below the computer keyboard, in a wheat field. & five sandwiches on a carpeted floor. \\
\checkmark & -- & -- & a photo of a frog and a pineapple, the frog is below the pineapple, on a wooden floor. & a photo of five goats, in a forest clearing. \\
\checkmark & -- & \checkmark & a photo of a globe and a penguin, the globe is below the penguin, in a pine forest. & a photo of five balloons, in a wheat field. \\
\checkmark & \checkmark & -- & an image showing a duck below a violin, on a sandy desert. & a realistic shot of five crowns, on a stone plaza. \\
\checkmark & \checkmark & \checkmark & a realistic shot of a hat with a camera below it, on a dark blue background. & five candles on a cobblestone street. \\
\bottomrule
\end{tabular}
\caption{\textbf{Held-out \conf prompts.} For ``below'' and ``5 objects'', the first prompt by ID in each of the 8 cells of 25 (\cref{app:confirm}): objects from a list outside the checker's COCO classes (new obj.), a new phrasing (new phr.), a new scene, or none of these (first row, the training templates).}
\label{fig:confirmset}
\end{figure*}

\paragraph{\contact.}
The set has 200 prompts: half say ``on top of'' and half use other contact verbs; all are scored by
the question ``Is the A on top of the B?''.

\paragraph{Replacements and freeze.}
Before the freeze we replaced 45 prompts naming shoes (counted as pairs), skis and scissors, whose
counts or positions the scorer misreads; the replacements were chosen with new seeds after looking
only at base-model and teacher images. %
Both sets, the scoring code and the comparisons were frozen
at 2026-11-13 18:23:45 Beijing time, before any student was scored on them. %

\section{Data acquisition cost}
\label{app:cost}

\begin{table}[t]
\centering
\caption{\textbf{GPU-hours to build each pool} (H100; \HO, added later, was not logged). Teacher:
drawing and checking the teacher images. Probe read: the student images this pool's rule reads; for \CQ, 0.47 of the 0.66 is measured for 576
teacher probe images and 0.19 is estimated for 576 reused student images. Full:
as if the pool alone paid for the whole shared probe batches it reads (21.0 GPU-hours; 25.6 for \PP
and \FO, which also read a second batch of 4.6). \SR's subtype failure rates come from student images that were
not logged, so its lower bound counts the teacher only. Training costs the same for all pools.}
\label{tab:cost}
\footnotesize
\setlength{\tabcolsep}{4pt}
\begin{tabular}{@{}l rrrr@{}}
\toprule
Pool & Teacher & Probe read & Lower bound & Full \\
\midrule
\ours & 6.39 & 1.40 & 7.79 & 27.43 \\ %
\FP & 7.24 & 1.40 & 8.64 & 28.28 \\
\PP & 5.76 & 1.78 & 7.53 & 31.37 \\
\SR & 6.42 & not logged & 6.42 & not logged \\
\RD & 5.38 & 0.00 & 5.38 & 5.38 \\
\CQ & 5.79 & 0.66 & 6.45 & 6.45 \\
\FO & 4.60 & 2.34 & 6.94 & 30.22 \\
\VS & 6.66 & 1.40 & 8.06 & 27.70 \\
\LR & 4.72 & 1.40 & 6.12 & 25.76 \\
\bottomrule
\end{tabular}
\end{table}

\section{Further readouts}
\label{app:readouts}

\begin{table}[t]
\centering
\caption{\textbf{Prompt budget sweep} on \conf (8 seeds). The $-$\RD and $-$\SR rows give \ours minus
that pool, with positive seeds in parentheses. Post hoc.}
\label{tab:scale_supp}
\footnotesize
\setlength{\tabcolsep}{2.5pt}
\begin{tabular}{@{}l rrrr@{}}
\toprule
Prompts & 125 & 250 & 500 & 1,000 \\
\midrule
\ours $\uparrow$ & 91.14 & 91.00 & 89.70 & 90.51 \\ %
$-$\RD & +2.27 (6) & +2.27 (8) & $-$0.27 (3) & +1.05 (6) \\
$-$\SR & +1.76 (6) & +0.58 (4) & $-$0.17 (4) & +0.62 (5) \\
\FP $\uparrow$ & 90.51 & 90.07 & 89.17 & 90.30 \\
\PP $\uparrow$ & 90.12 & 88.56 & 90.02 & 90.28 \\
\bottomrule
\end{tabular}
\end{table}

\begin{table}[t]
\centering
\caption{\textbf{Classic GenEval 6-task breakdown} (SD3.5-Medium, 3 of the 8 seeds; Mask2Former +
ViT-L-14; 553 prompts $\times$ 4 images).}
\label{tab:geneval_supp}
\scriptsize
\setlength{\tabcolsep}{1.8pt}
\begin{tabular}{@{}l rr rrrrrr@{}}
\toprule
Model & All $\uparrow$ & P+C $\uparrow$ & 1-Obj & 2-Obj & Cnt. & Col. & Pos. & Attr. \\
\midrule
Base & 68.27 & 43.78 & 99.06 & 84.34 & 62.81 & 81.65 & 24.75 & 57.00 \\ %
Teacher & 93.05 & 94.81 & 100.00 & 98.48 & 93.12 & 86.97 & 96.50 & 83.25 \\
\midrule
\ours, 1st & 88.02 & 80.28 & 99.38 & \textbf{98.74} & 90.31 & 88.03 & 70.25 & 81.42 \\
\ours, ver. & \textbf{89.72} & 84.43 & \textbf{100.00} & 98.40 & \textbf{93.02} & \textbf{88.56} & 75.83 & \textbf{82.50} \\
\RD & 89.38 & \textbf{84.54} & 99.79 & 98.48 & 92.92 & 86.44 & 76.17 & \textbf{82.50} \\
\HO & 88.73 & 82.07 & 99.79 & 98.32 & 90.31 & 88.39 & 73.83 & 81.75 \\
\CQ & 87.68 & 84.32 & 99.90 & 97.81 & 91.56 & 82.62 & \textbf{77.08} & 77.08 \\
\bottomrule
\end{tabular}
\end{table}

\paragraph{Verification control on other sets (\FP, 8 seeds).}
Verified minus first draw: \idset 91.44 \vs 86.31 (+5.14, 8/8); GenEval\,2 23.75 \vs 21.84
(+1.90, 8/8); GenEval\,2 ``on top'' +1.28 (7/8); \contact +1.36 (6/8). %

\paragraph{Full public-benchmark readouts (GenEval\,2, classic GenEval and T2I-CompBench).}
All saved step-2,000 checkpoints were evaluated without retraining at 0\% missing scores and $\le 0.05$
parity drift on re-scored base models (Tabs.~2--4 and Fig.~5 of the
main paper): %
\begin{itemize}[leftmargin=*,itemsep=1pt,topsep=2pt]
\item \emph{SD3.5-Medium on GenEval\,2 (8 seeds, 800 prompts, Soft-TIFA GM / practiced atoms / 104
  ``on top of'' atoms):} base 18.19 / 47.67 / 45.34, Flow-GRPO teacher 24.50 / 56.06 / 28.11, \ours
  verified 23.29 / 56.17 / 25.82, \ours first-draw 22.25 / 54.66 / 25.72 ($\Delta=+1.04$, 6/8, $t$
  $[-0.60,+2.68]$; practiced $+1.52$, 8/8), \FP verified 23.75 / 56.37 / 25.77, \FP first-draw 21.84 /
  55.17 / 24.49 ($\Delta=+1.90$, 8/8, $t$ $[+0.85,+2.96]$), \RD 22.57 / 55.96 / 22.97, \HO 22.59 /
  55.49 / 27.40, \CQ 21.04 / 53.57 / 24.47 (\ours $-\,\CQ=+2.25$, 7/8, $t$ $[+0.77,+3.74]$; practiced
  $+2.61$, 8/8), \SR 22.26, \PP 23.74, \FO 24.14, \LR 23.79, \VS 23.32. On the 3-seed contact cells,
  the split verifier scores 29.50 on ``on top of'' \vs 25.45 for the box checker ($+4.05$, 3/3).
\item \emph{Sana-1.6B on GenEval\,2 (3 of the 8 seeds):} sana-base 13.71 / 43.47 / 36.90, \ours verified
  16.79 / 47.57 / 23.45, \ours first-draw 16.65 / 46.93 / 25.42 ($\Delta=+0.14$, 1/3; practiced
  $+0.64$), \RD 16.89, \CQ 16.03 (\ours $-\,\CQ=+0.76$, 2/3); on ``on top of'' the contact cells score 27.53 with
  the split verifier \vs 25.45 with the box checker ($+2.08$, 2/3).
\item \emph{SD3.5-Medium on classic GenEval (3 of the 8 seeds, 6-task breakdown in \cref{tab:geneval_supp}):}
  \ours verified beats first-draw by $+1.70$ Overall and $+4.15$ on Position+Counting (2/3 seeds; Pos.\
  $+5.58$, Cnt.\ $+2.71$), beats \HO by $+0.99$ Overall and $+2.35$ on Position+Counting (3/3), ties \RD
  ($+0.34$ Overall, $-$0.11 Pos+Cnt), and beats \CQ by $+2.04$ Overall (3/3, driven by \CQ's 24 \vs 185
  color prompts) while tying on Position+Counting ($+0.10$, 2/3, as classic GenEval tests only 2--4
  counts).
\item \emph{T2I-CompBench official UniDet (3 of the 8 seeds, Avg / Spatial / Numeracy):} SD3.5-M base
  43.82 (29.34 / 58.31), teacher 61.70 (53.02 / 70.39), \ours verified 59.63 (49.59 / 69.66), \ours
  first-draw 58.83 (48.96 / 68.70; $\Delta=+0.80$, 2/3; practiced 2D/$\le 5$-count subset $+1.02$),
  \FP verified 60.17 (50.76 / 69.57), \FP first-draw 59.35 (49.76 / 68.94; $\Delta=+0.82$, 2/3;
  practiced $+1.19$, 3/3), \RD 59.68 (50.91 / 68.45), \HO 59.82 (50.26 / 69.38), \CQ 59.80 (50.93 /
  68.68), \SR 60.00 (51.13 / 68.87), \LR 59.50 (49.57 / 69.42). Sana-1.6B base 45.06 (30.18 / 59.93),
  \ours verified 56.76 (48.04 / 65.49), \ours first-draw 55.78 (45.98 / 65.58; $\Delta=+0.98$, 3/3;
  practiced $+1.16$, 3/3), \RD 56.71 (47.89 / 65.52), \CQ 57.24 (48.53 / 65.95).
\end{itemize}

\paragraph{DrawBench (\FP, 3 seeds).}
ImageReward / PickScore / diversity: base 0.846 / 22.420 / 0.01439, teacher 1.059 / 22.533 /
0.01080, \FP 1.224 / 22.571 / 0.00852, \RD 1.176 / 22.538 / 0.00880. \FP minus teacher: ImageReward
+0.165 (3/3), diversity $-$0.00228 (0/3). %

\begin{figure*}[p]
\centering
\newcommand{\supimg}[3]{%
  \begin{tikzpicture}[baseline=(img.south)]
    \node[inner sep=0pt] (img) {#1};
    \node[circle, fill=#2, text=white, inner sep=0.4pt, font=\tiny\bfseries]
      at ($(img.north east)+(-0.11,-0.11)$) {#3};
  \end{tikzpicture}}
\def\supw{0.111\linewidth}
\newcommand{\supp}[1]{\multicolumn{4}{c}{\parbox[t]{0.448\linewidth}{\centering\tiny #1}}}
\setlength{\tabcolsep}{0.6pt}
\renewcommand{\arraystretch}{0.5}
\begin{tabular}{@{}c cccc@{\hspace{6pt}}cccc@{}}
 & \scriptsize Teacher & \scriptsize 1st draw & \scriptsize Verified & \scriptsize \RD & \scriptsize Teacher & \scriptsize 1st draw & \scriptsize Verified & \scriptsize \RD \\[2pt]
\rotatebox{90}{\scriptsize\ below} & \supimg{\includegraphics[width=\supw]{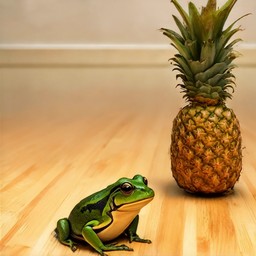}}{oiGreen!85!black}{\checkmark} & \supimg{\includegraphics[width=\supw]{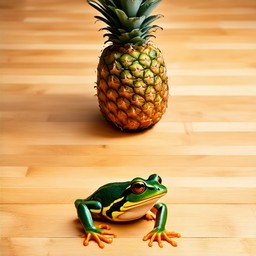}}{oiGreen!85!black}{\checkmark} & \supimg{\includegraphics[width=\supw]{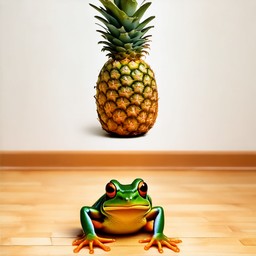}}{oiGreen!85!black}{\checkmark} & \supimg{\includegraphics[width=\supw]{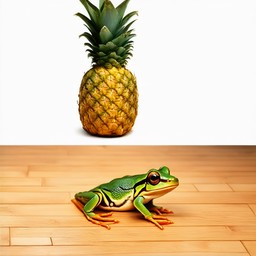}}{oiGreen!85!black}{\checkmark} & \supimg{\includegraphics[width=\supw]{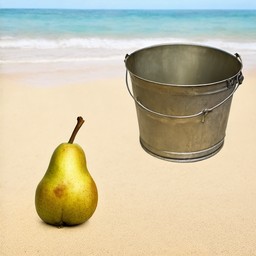}}{oiGreen!85!black}{\checkmark} & \supimg{\includegraphics[width=\supw]{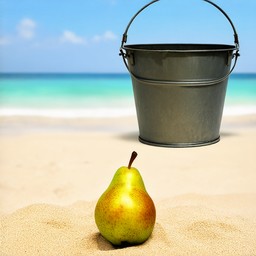}}{oiGreen!85!black}{\checkmark} & \supimg{\includegraphics[width=\supw]{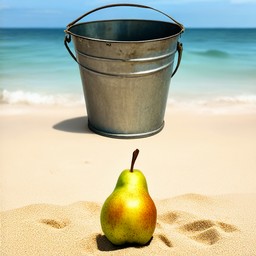}}{oiGreen!85!black}{\checkmark} & \supimg{\includegraphics[width=\supw]{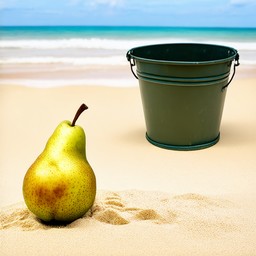}}{oiGreen!85!black}{\checkmark} \\[1pt]
 & \supp{a frog and a pineapple, the frog is below the pineapple, on a wooden floor} & \supp{a pear and a bucket, the pear is below the bucket, on a sandy beach} \\[4pt]
\rotatebox{90}{\scriptsize\ above} & \supimg{\includegraphics[width=\supw]{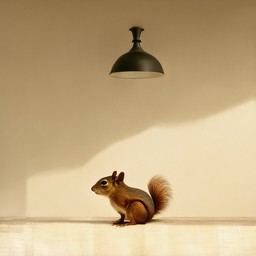}}{oiGreen!85!black}{\checkmark} & \supimg{\includegraphics[width=\supw]{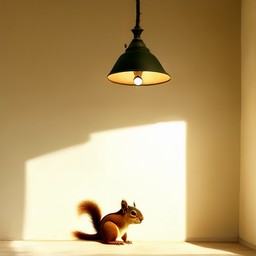}}{oiGreen!85!black}{\checkmark} & \supimg{\includegraphics[width=\supw]{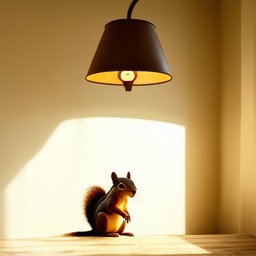}}{oiGreen!85!black}{\checkmark} & \supimg{\includegraphics[width=\supw]{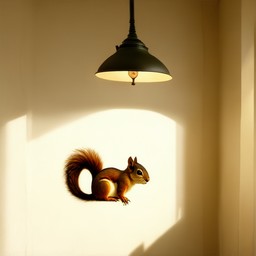}}{oiGreen!85!black}{\checkmark} & \supimg{\includegraphics[width=\supw]{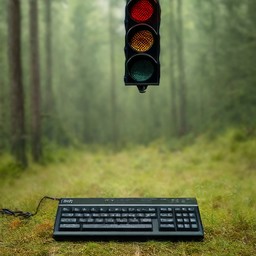}}{oiGreen!85!black}{\checkmark} & \supimg{\includegraphics[width=\supw]{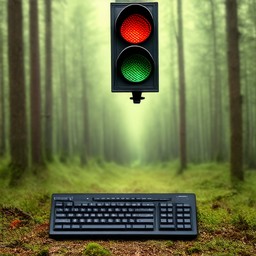}}{oiGreen!85!black}{\checkmark} & \supimg{\includegraphics[width=\supw]{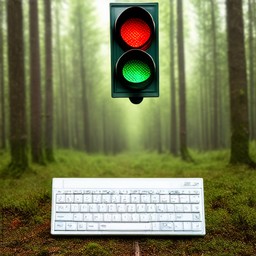}}{oiGreen!85!black}{\checkmark} & \supimg{\includegraphics[width=\supw]{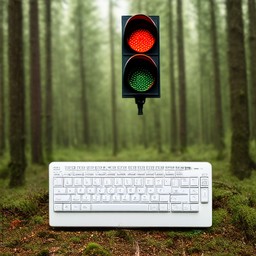}}{oiGreen!85!black}{\checkmark} \\[1pt]
 & \supp{a lamp above a squirrel, in a sunlit room with bare walls} & \supp{a traffic light and a computer keyboard, the traffic light is above the computer keyboard, in a forest clearing} \\[4pt]
\rotatebox{90}{\scriptsize\ left of} & \supimg{\includegraphics[width=\supw]{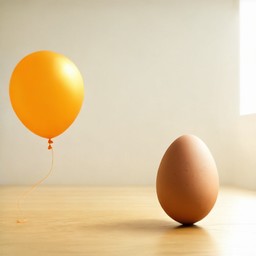}}{oiGreen!85!black}{\checkmark} & \supimg{\includegraphics[width=\supw]{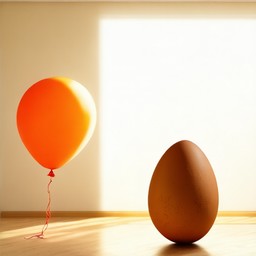}}{oiGreen!85!black}{\checkmark} & \supimg{\includegraphics[width=\supw]{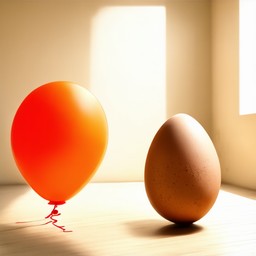}}{oiGreen!85!black}{\checkmark} & \supimg{\includegraphics[width=\supw]{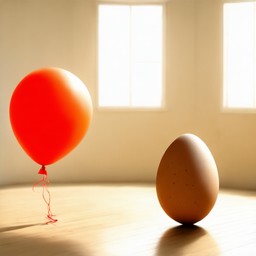}}{oiGreen!85!black}{\checkmark} & \supimg{\includegraphics[width=\supw]{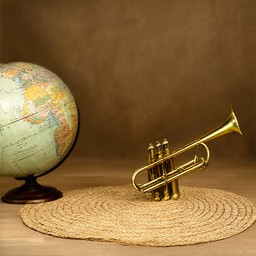}}{oiGreen!85!black}{\checkmark} & \supimg{\includegraphics[width=\supw]{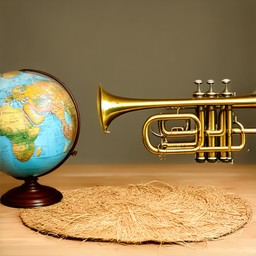}}{oiGreen!85!black}{\checkmark} & \supimg{\includegraphics[width=\supw]{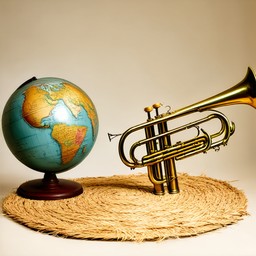}}{oiGreen!85!black}{\checkmark} & \supimg{\includegraphics[width=\supw]{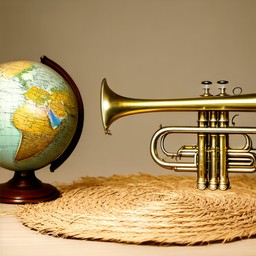}}{oiGreen!85!black}{\checkmark} \\[1pt]
 & \supp{a picture where the balloon is to the left of the egg, in a sunlit room with bare walls} & \supp{a globe to the left of a trumpet, on a straw mat} \\[4pt]
\rotatebox{90}{\scriptsize\ right of} & \supimg{\includegraphics[width=\supw]{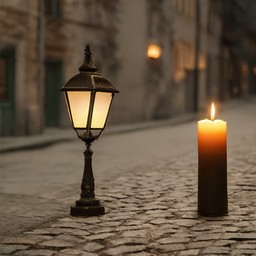}}{oiGreen!85!black}{\checkmark} & \supimg{\includegraphics[width=\supw]{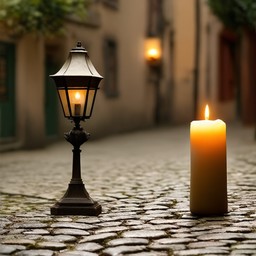}}{oiGreen!85!black}{\checkmark} & \supimg{\includegraphics[width=\supw]{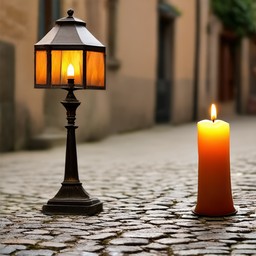}}{oiGreen!85!black}{\checkmark} & \supimg{\includegraphics[width=\supw]{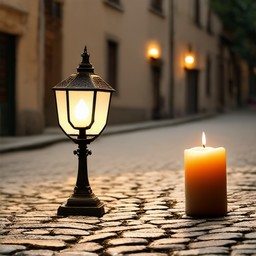}}{oiGreen!85!black}{\checkmark} & \supimg{\includegraphics[width=\supw]{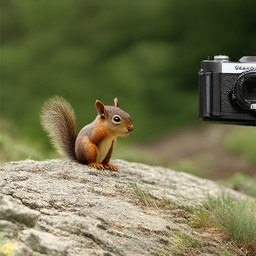}}{oiGreen!85!black}{\checkmark} & \supimg{\includegraphics[width=\supw]{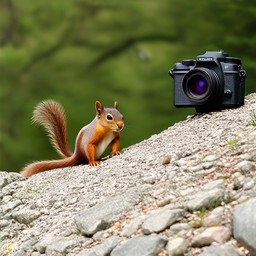}}{oiGreen!85!black}{\checkmark} & \supimg{\includegraphics[width=\supw]{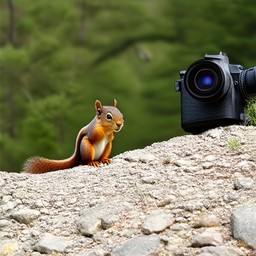}}{oiGreen!85!black}{\checkmark} & \supimg{\includegraphics[width=\supw]{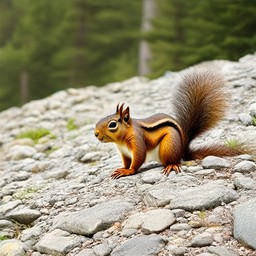}}{oiOrange!90!black}{$\times$} \\[1pt]
 & \supp{a candle and a lamp, the candle is to the right of the lamp, on a cobblestone street} & \supp{a picture where the camera is to the right of the squirrel, on a rocky hillside} \\[4pt]
\rotatebox{90}{\scriptsize\ 3 objects} & \supimg{\includegraphics[width=\supw]{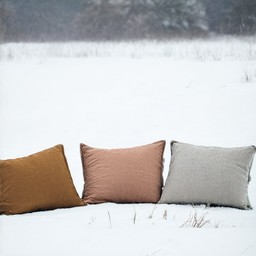}}{oiGreen!85!black}{\checkmark} & \supimg{\includegraphics[width=\supw]{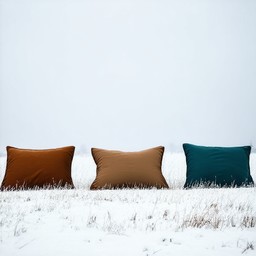}}{oiGreen!85!black}{\checkmark} & \supimg{\includegraphics[width=\supw]{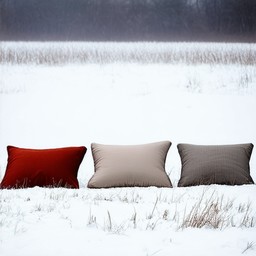}}{oiGreen!85!black}{\checkmark} & \supimg{\includegraphics[width=\supw]{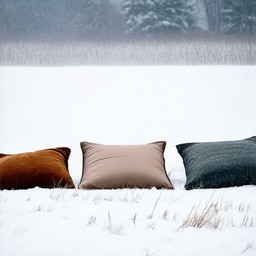}}{oiGreen!85!black}{\checkmark} & \supimg{\includegraphics[width=\supw]{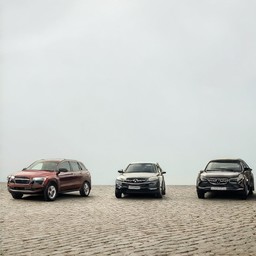}}{oiGreen!85!black}{\checkmark} & \supimg{\includegraphics[width=\supw]{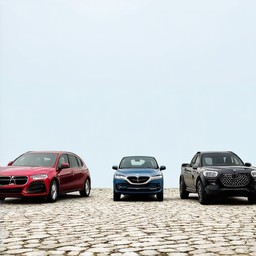}}{oiGreen!85!black}{\checkmark} & \supimg{\includegraphics[width=\supw]{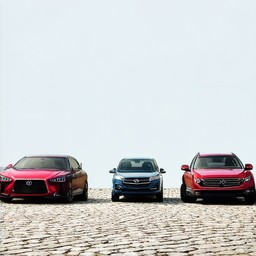}}{oiGreen!85!black}{\checkmark} & \supimg{\includegraphics[width=\supw]{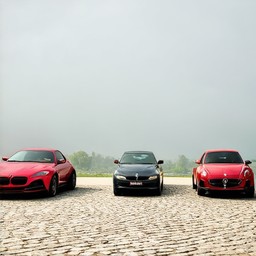}}{oiGreen!85!black}{\checkmark} \\[1pt]
 & \supp{three pillows, in a snowy field} & \supp{a picture with three cars in view, on a stone plaza} \\[4pt]
\rotatebox{90}{\scriptsize\ 4 objects} & \supimg{\includegraphics[width=\supw]{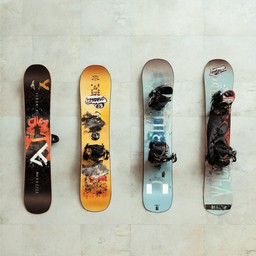}}{oiGreen!85!black}{\checkmark} & \supimg{\includegraphics[width=\supw]{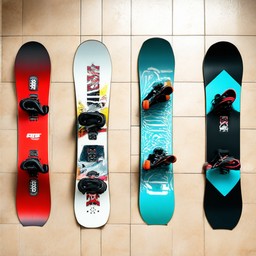}}{oiGreen!85!black}{\checkmark} & \supimg{\includegraphics[width=\supw]{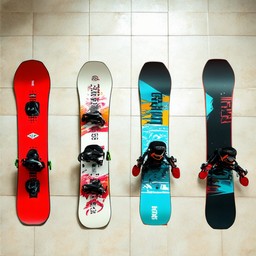}}{oiGreen!85!black}{\checkmark} & \supimg{\includegraphics[width=\supw]{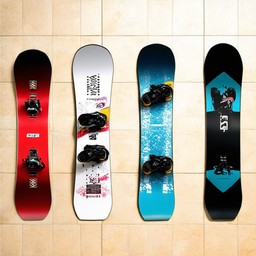}}{oiGreen!85!black}{\checkmark} & \supimg{\includegraphics[width=\supw]{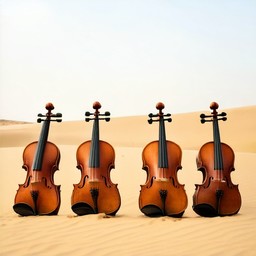}}{oiGreen!85!black}{\checkmark} & \supimg{\includegraphics[width=\supw]{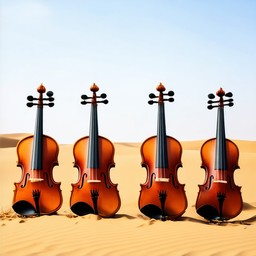}}{oiGreen!85!black}{\checkmark} & \supimg{\includegraphics[width=\supw]{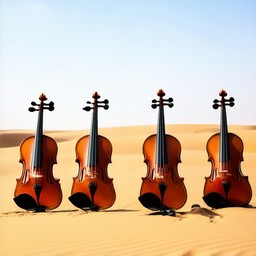}}{oiGreen!85!black}{\checkmark} & \supimg{\includegraphics[width=\supw]{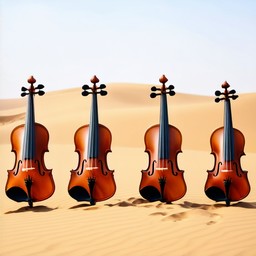}}{oiGreen!85!black}{\checkmark} \\[1pt]
 & \supp{a realistic shot of four snowboards, on a tiled floor} & \supp{four violins, on a sandy desert} \\[4pt]
\rotatebox{90}{\scriptsize\ 5 objects} & \supimg{\includegraphics[width=\supw]{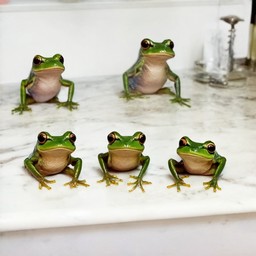}}{oiGreen!85!black}{\checkmark} & \supimg{\includegraphics[width=\supw]{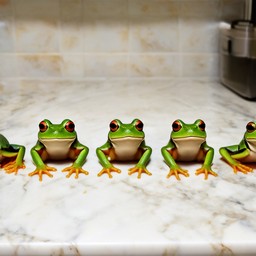}}{oiGreen!85!black}{\checkmark} & \supimg{\includegraphics[width=\supw]{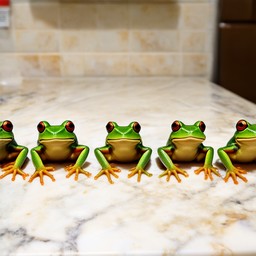}}{oiGreen!85!black}{\checkmark} & \supimg{\includegraphics[width=\supw]{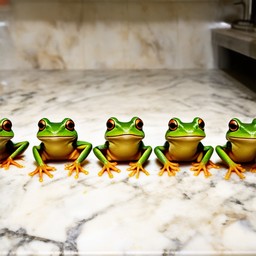}}{oiGreen!85!black}{\checkmark} & \supimg{\includegraphics[width=\supw]{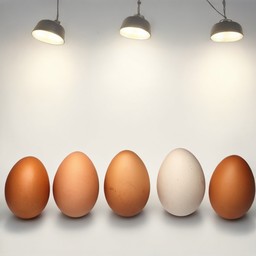}}{oiGreen!85!black}{\checkmark} & \supimg{\includegraphics[width=\supw]{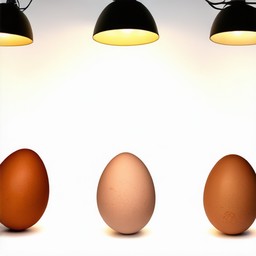}}{oiOrange!90!black}{$\times$} & \supimg{\includegraphics[width=\supw]{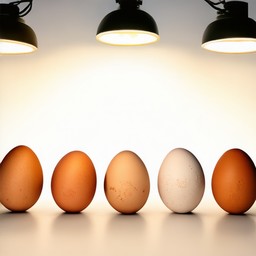}}{oiGreen!85!black}{\checkmark} & \supimg{\includegraphics[width=\supw]{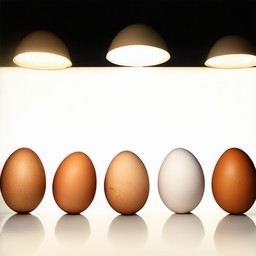}}{oiGreen!85!black}{\checkmark} \\[1pt]
 & \supp{five frogs, on a marble countertop} & \supp{five eggs, under bright studio lights} \\[4pt]
\end{tabular}
\caption{\textbf{Uncurated \conf samples} from the second-judge sample (models of training seed 2111277986; one image per prompt, with the same image index for every model). Per subtype, the first two prompts of the sample by ID, chosen before any image was viewed. Columns: the teacher's single draw, \FP trained on first-draw or on verified targets, and \RD (verified). Badges give the Qwen3-VL judge's answer to the target question ({\color{oiGreen!85!black}\checkmark}: probability $\ge 0.5$).}
\label{fig:confirm}
\end{figure*}

\paragraph{Second judge (\FP, seed 2111277986, 200+40 prompts, 1 image).}
The sample was fixed before any image was viewed: 200 \conf prompts stratified by subtype and 40
\contact prompts, one image index per prompt, the same for every model. Gemini answered every question
Qwen3-VL was scored on and whether the image meets the whole prompt; 3 of the 920 replies could not
be parsed and are left out. Rates are weighted by subtype size, and differences are paired over the
prompts both models have (198 for \FP minus teacher). %
Qwen target / Gemini target / Gemini whole prompt: teacher 90.0 / 87.6 / 84.6, \FP 92.4 / 91.0 /
90.5, \RD 92.8 / 91.1 / 89.6, \FP with first draws 86.6 / 85.6 / 82.7. %
\cref{fig:confirm} shows the images of the first two \conf prompts of this sample in each subtype.

\begin{figure*}[t]
\centering
\newcommand{\supimg}[3]{%
  \begin{tikzpicture}[baseline=(img.south)]
    \node[inner sep=0pt] (img) {#1};
    \node[circle, fill=#2, text=white, inner sep=0.4pt, font=\tiny\bfseries]
      at ($(img.north east)+(-0.11,-0.11)$) {#3};
  \end{tikzpicture}}
\def\supw{0.148\linewidth}
\newcommand{\supl}[1]{\parbox[t]{\supw}{\centering\tiny #1}}
\setlength{\tabcolsep}{0.7pt}
\renewcommand{\arraystretch}{0.5}
\begin{tabular}{@{}c cccccc@{}}
\rotatebox{90}{\scriptsize\ 512 px} & \supimg{\includegraphics[width=\supw]{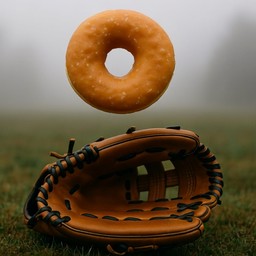}}{oiGreen!85!black}{\checkmark} & \supimg{\includegraphics[width=\supw]{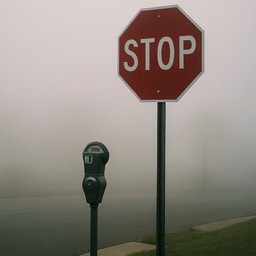}}{oiOrange!90!black}{$\times$} & \supimg{\includegraphics[width=\supw]{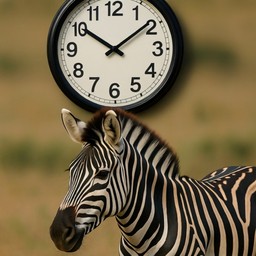}}{oiGreen!85!black}{\checkmark} & \supimg{\includegraphics[width=\supw]{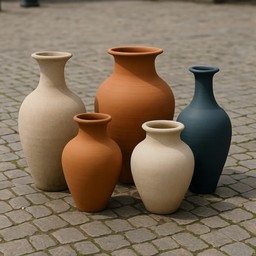}}{oiGreen!85!black}{\checkmark} & \supimg{\includegraphics[width=\supw]{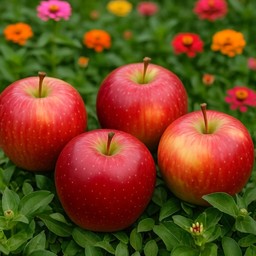}}{gray!70}{?} & \supimg{\includegraphics[width=\supw]{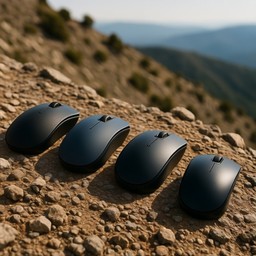}}{oiGreen!85!black}{\checkmark} \\[1pt]
\rotatebox{90}{\scriptsize\ 1024 px} & \supimg{\includegraphics[width=\supw]{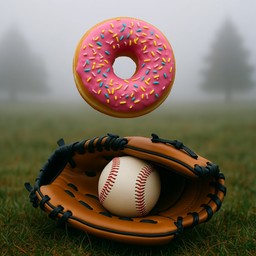}}{oiGreen!85!black}{\checkmark} & \supimg{\includegraphics[width=\supw]{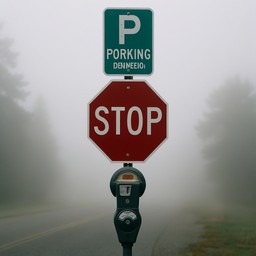}}{oiOrange!90!black}{$\times$} & \supimg{\includegraphics[width=\supw]{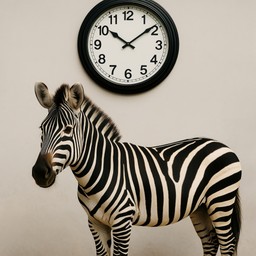}}{oiGreen!85!black}{\checkmark} & \supimg{\includegraphics[width=\supw]{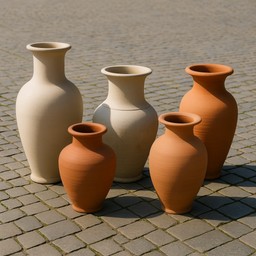}}{oiGreen!85!black}{\checkmark} & \supimg{\includegraphics[width=\supw]{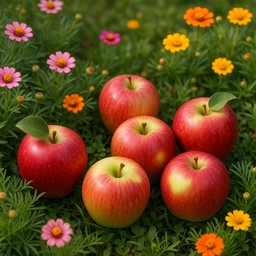}}{gray!70}{?} & \supimg{\includegraphics[width=\supw]{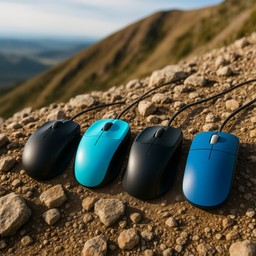}}{oiGreen!85!black}{\checkmark} \\[1pt]
 & \supl{a donut and a baseball glove, the donut is above the baseball glove, in a foggy landscape} & \supl{a stop sign and a parking meter, the stop sign is below the parking meter, in a foggy landscape} & \supl{a zebra and a clock, the zebra is below the clock} & \supl{five vases, on a stone plaza} & \supl{five apples, in a flower garden} & \supl{four computer mice, on a rocky hillside} \\
\end{tabular}
\caption{\textbf{Qwen-Image's first draws at 512\,px and at 1024\,px} for the same \ours prompts: the first three position and the first three count prompts by ID, chosen before any image was viewed. Both rows are shown at the same size (1024\,px draws were downsampled to 512\,px for training, \cref{app:readouts}). Badges give the box checker's verdict on each draw ({\color{gray!70}?}: uncertain).}
\label{fig:qwen}
\end{figure*}

\paragraph{Teacher from another model family (Qwen-Image, 512\,px and 1024$\to$512\,px, \ours pool, 3 seeds).}
Of the 1,000 prompts of \ours, Qwen-Image at 512\,px produced a checker-accepted image within 8, 8 and 16 draws
for 928 prompts; the remaining 72 kept Qwen-Image's first draw in both training sets. On \conf (seeds
2111277986 / 167930152 / 249530483), Qwen-Image at 512\,px scores 86.14 (position 90.87, count 81.41, \contact
98.96); the verified student scores 83.85 (position 91.38, count 76.33, \contact 96.65) and the
first-draw student 78.35 (position 88.52, count 68.18, \contact 95.27), giving paired differences of
+5.41, +3.94 and +7.16 (+5.50 mean, $t$ interval $[+1.50,+9.51]$, prompt interval $[+4.69,+6.35]$).
By subtype (verified student minus Qwen-Image): below +7.87, above +1.93, left of $-$2.73, right of
$-$5.05; 3 objects $-$9.33, 4 objects $-$8.45, 5 objects +2.53. %
Because Qwen-Image is natively trained above 1024\,px and produces crowded 512\,px layouts, we also
redrew \ours at 1024\,px, kept the 933 prompts with a checker-accepted image in both training sets (76.8\%
already passing on the first draw), and downsampled targets to 512\,px for the same 2,000-step SFT:
the verified 512\,px student rises to 87.58 \vs 85.18 for first draws (+2.39 across 3/3 seeds:
+1.07 / +3.00 / +3.11; position 92.43 \vs 90.58, count 82.72 \vs 79.79; \contact 95.89 \vs 95.52;
intermediate checkpoints at steps 500 / 1,000 / 1,500 / 2,000 score 81.07 / 85.16 / 86.79 / 87.58,
EMA 87.43). This exceeds Qwen-Image at 512\,px (86.14), while Qwen-Image evaluated directly at
1024\,px reaches 91.48 (note that the judge views 1024\,px teacher images \vs 512\,px student
images). %
\cref{fig:qwen} shows Qwen-Image's first draws at both resolutions.

\paragraph{Image source $\times$ verifier and false-pass dose, per seed and on other sets.}
Seeds 2111277986 / 167930152 / 249530483 on \contact: teacher and box checker 85.25 / 83.68 / 83.61;
teacher and split verifier 91.22 / 89.13 / 89.34; base and split verifier 95.93 / 93.61 / 95.90;
base and box checker 95.79 / 94.44 / 95.69. On \conf the four cells score 91.09, 90.84, 90.42 and
90.73; on GenEval\,2 ``on top of'' (104 atoms) the teacher cells score 25.45 (box) and 29.50
(split, $+4.05$, 3/3, $t\,[+1.83,+6.26]$). %
Swapping 33, 65 or 98 of the split verifier's 119 teacher ``on top of'' images (28\%, 55\%, 82\%, out
of 98 prompts that also have a box-accepted floating teacher image) gives \contact 87.06 / 82.35 /
82.90 (mean 84.10, $-$5.80 \vs 0\%), 86.66 / 78.42 / 80.34 (81.80, $-$8.09), and 79.41 / 77.63 / 71.19
(76.07, $-$13.82); \conf means are 91.30, 91.07 and 89.81. %

\paragraph{Second open-source judge (InternVL3.5-38B, 3 seeds).}
OpenGVLab InternVL3.5-38B (InternViT-6B-448px-V2.5 + Qwen3-32B, non-thinking
Soft-TIFA) re-scored saved confirmation images on seeds 2111277986 / 167930152 / 249530483 (0\%
missing; 95.90\% binary agreement at 0.5 and Pearson $r=0.892$ with Qwen3-VL-8B over 101,600 target
questions). On \conf: base 58.22 (pos.\ 71.65, cnt.\ 44.79), Flow-GRPO teacher 83.86 (88.13 / 79.58),
\ours verified 89.48 (91.10 / 87.86; seeds 89.18 / 89.24 / 90.01), \ours first-draw 84.08 (88.37 /
79.80; seeds 84.00 / 83.19 / 85.05; $\Delta=+5.40$, 3/3, $t$ $[+3.95,+6.84]$), \RD 87.74 (seeds 89.49 /
87.89 / 85.83; \ours $-\,\RD=+1.74$, 2/3, $t$ $[-3.90,+7.39]$), \HO 85.13 (seeds 88.09 / 85.29 / 82.00;
\ours $-\,\HO=+4.35$, 3/3), and \CQ 84.36 (seeds 84.95 / 86.32 / 81.81; \ours $-\,\CQ=+5.12$, 3/3). On \contact: base
94.33, teacher 79.68, box checker 86.25 (87.06 / 85.89 / 85.82), split verifier 91.47 (93.05 / 90.43 /
90.94; $\Delta=+5.22$, 3/3, $t$ $[+3.41,+7.02]$). %

\paragraph{Heterogeneous student backbone (Sana-1.6B-512px, 3 seeds).}
We fine-tuned Sana-1.6B-512px (Linear-DiT, 32$\times$ DC-AE, Gemma-2-2B; LoRA rank 32,
$\alpha=64$ on self- and cross-attention projections, 20-step sampling at guidance 4.5) with the exact
same 2,000-step recipe and Flow-GRPO teacher image sets on seeds 2111277986 / 167930152 / 249530483
(0\% missing). Un-finetuned Sana-1.6B scores 58.42 on \conf (pos.\ 64.90, cnt.\ 51.94) and 89.88 on
\contact. After SFT: \ours verified scores 88.10 (pos.\ 89.39, cnt.\ 86.80, \contact 77.22; seeds
87.52 / 87.82 / 88.94, beating the Flow-GRPO teacher's 85.43 in 3/3 seeds), \ours first-draw 83.74
(86.57 / 80.91, \contact 76.71; seeds 81.53 / 83.88 / 85.82; $\Delta=+4.36$, 3/3, $t$ $[+0.68,+8.03]$),
\RD 88.32 (88.47 / 88.17, \contact 78.51; seeds 87.53 / 90.15 / 87.28; \ours $-\,\RD=-0.22$, 1/3, $t$
$[-5.19,+4.75]$), \CQ 84.64 (86.56 / 82.71, \contact 78.15; seeds 84.99 / 84.21 / 84.70;
\ours $-\,\CQ=+3.46$, 3/3, $t$ $[+1.31,+5.62]$); for the contact cells, box checker scores 75.25 on \contact
(\conf 88.49) and split verifier 82.55 (\conf 88.49), giving $\Delta=+7.30$ (3/3: +6.80 / +6.15 /
+8.96, $t$ $[+3.65,+10.95]$). %

\section{Image source \texorpdfstring{$\times$}{x} verifier: construction}
\label{app:split}

\paragraph{Prompts and images.}
The 198 training prompts read ``a photo of an A and a B, the A is on top of the B.'', with A a small
object and B a support it can rest on, from a fixed list of pairs; no test prompt names the two
objects of a pair together.
The box checker reads them as ``A above B''. For each prompt the cells use the teacher's first 8
images and up to 8 images of the base model, in drawing order. %

\paragraph{Verifiers.}
The box checker accepts an image when every constraint passes (Sec.~3 of the main paper).
The split verifier drops the box rule: the detector must find both objects, and Gemini
(gemini-3.8-flash), given the image and the two object names, must answer that A rests on top of B,
that they touch, that A does not float and that they are not fused. Its instruction asks, in
JSON, whether A and B are visible and how many of each there are, and: ``does A appear higher in the
picture than B, from the viewer's point of view?'', ``does A physically rest on top of B, held up by
B's upper surface?'', ``do A and B touch?'', ``does A hang in the air with nothing holding it up?''
and whether A and B are ``merged into one hybrid object''. %

\paragraph{Cells.}
Each cell takes, per prompt, the first image its verifier accepts. Teacher and box checker found one
for 198 prompts, teacher and split verifier for 141, base model and split verifier for 193, base
model and box checker for 171. The 119 prompts with an image in every cell are used in all four
cells: they replace 119 of the 368 position-failure prompts of the \FP pool (of its 419 position-template
prompts, 51 were retained for an object or color failure), in proportion to its relations
(below 71, above 19, left of 5, right of 24) and in a seeded order, so the four pools differ only in
these images. The two verifiers chose different teacher images for 60 of the 119 prompts. Before the
build we fixed three stop rules, none of which fired: fewer than 100 prompts; the detector missing an
object in more than 20\% of the images Gemini finds correct (teacher 4.2\%, base model 19.7\%); more
than 5\% failed Gemini calls (1 of 3,168). %

\section{Compound prompts: false passes for ``below''}
\label{app:below}

\begin{figure*}[t]
\centering
\newcommand{\supimg}[3]{%
  \begin{tikzpicture}[baseline=(img.south)]
    \node[inner sep=0pt] (img) {#1};
    \node[circle, fill=#2, text=white, inner sep=0.4pt, font=\tiny\bfseries]
      at ($(img.north east)+(-0.11,-0.11)$) {#3};
  \end{tikzpicture}}
\def\supw{0.148\linewidth}
\newcommand{\supl}[1]{\parbox[t]{\supw}{\centering\tiny #1}}
\setlength{\tabcolsep}{0.7pt}
\renewcommand{\arraystretch}{0.5}
\begin{tabular}{@{}c cccccc@{}}
\multicolumn{7}{@{}l}{\scriptsize (a) Box-accepted teacher targets of the $2\times2$, ``A on top of B''} \\[2pt]
\rotatebox{90}{\tiny\ Gemini: floating} & \supimg{\includegraphics[width=\supw]{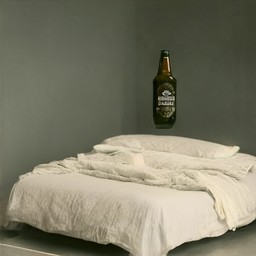}}{oiOrange!90!black}{$\times$} & \supimg{\includegraphics[width=\supw]{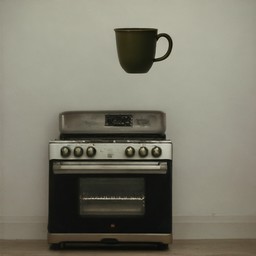}}{oiOrange!90!black}{$\times$} & \supimg{\includegraphics[width=\supw]{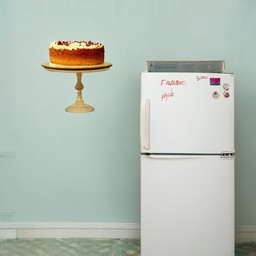}}{oiOrange!90!black}{$\times$} & \supimg{\includegraphics[width=\supw]{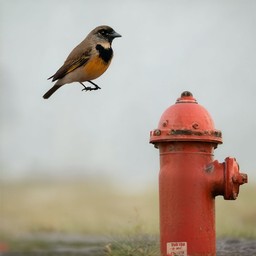}}{oiOrange!90!black}{$\times$} & \supimg{\includegraphics[width=\supw]{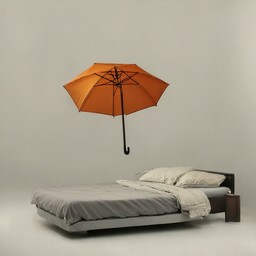}}{oiOrange!90!black}{$\times$} & \supimg{\includegraphics[width=\supw]{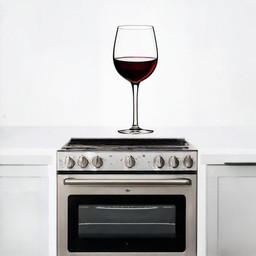}}{oiOrange!90!black}{$\times$} \\[1pt]
 & \supl{bottle on bed} & \supl{cup on oven} & \supl{cake on refrigerator} & \supl{bird on fire hydrant} & \supl{umbrella on bed} & \supl{wine glass on oven} \\[6pt]
\rotatebox{90}{\tiny\ not floating} & \supimg{\includegraphics[width=\supw]{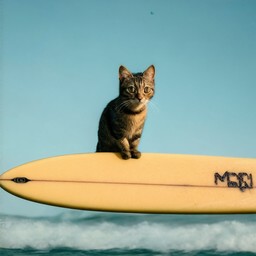}}{oiGreen!85!black}{\checkmark} & \supimg{\includegraphics[width=\supw]{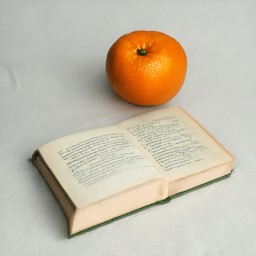}}{oiOrange!90!black}{$\times$} & \supimg{\includegraphics[width=\supw]{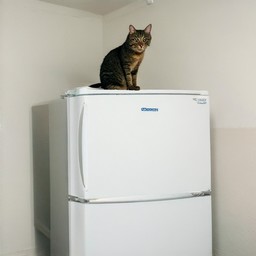}}{oiGreen!85!black}{\checkmark} & \supimg{\includegraphics[width=\supw]{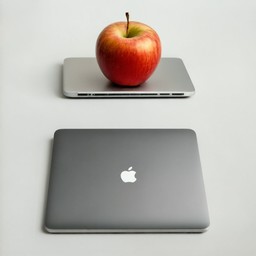}}{oiGreen!85!black}{\checkmark} & \supimg{\includegraphics[width=\supw]{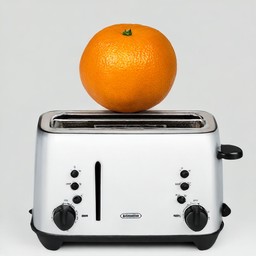}}{oiGreen!85!black}{\checkmark} & \supimg{\includegraphics[width=\supw]{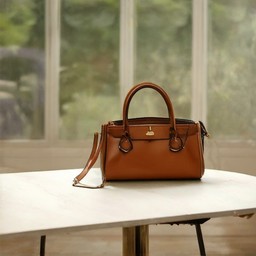}}{oiGreen!85!black}{\checkmark} \\[1pt]
 & \supl{cat on surfboard} & \supl{orange on book} & \supl{cat on refrigerator} & \supl{apple on laptop} & \supl{orange on toaster} & \supl{handbag on dining table} \\[6pt]
\multicolumn{7}{@{}l}{\scriptsize (b) Checker-passed compound ``below'' teacher images} \\[2pt]
\rotatebox{90}{\tiny\ Gemini: wrong} & \supimg{\includegraphics[width=\supw]{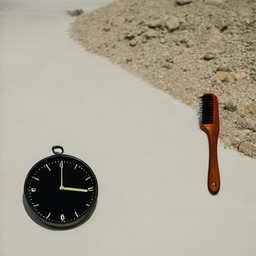}}{oiOrange!90!black}{$\times$} & \supimg{\includegraphics[width=\supw]{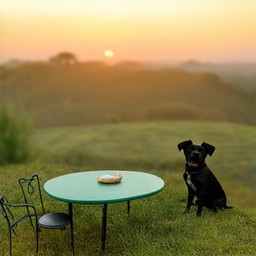}}{oiOrange!90!black}{$\times$} & \supimg{\includegraphics[width=\supw]{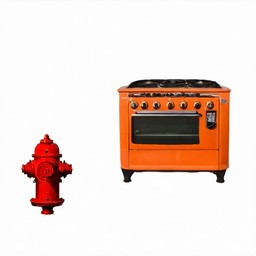}}{oiOrange!90!black}{$\times$} & \supimg{\includegraphics[width=\supw]{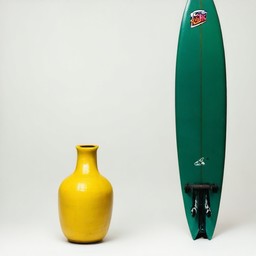}}{oiOrange!90!black}{$\times$} & \supimg{\includegraphics[width=\supw]{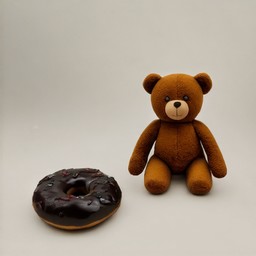}}{oiOrange!90!black}{$\times$} & \supimg{\includegraphics[width=\supw]{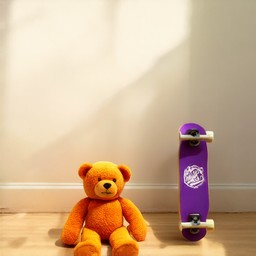}}{oiOrange!90!black}{$\times$} \\[1pt]
 & \supl{a black clock and a brown toothbrush, the clock is below the toothbrush, on a rocky hillside} & \supl{a green dining table and a black dog, the dining table is below the dog, at sunset on an open hill} & \supl{a red fire hydrant and an orange oven, the fire hydrant is below the oven, on a plain white background} & \supl{a yellow vase and a green surfboard, the vase is below the surfboard} & \supl{a black donut and a brown teddy bear, the donut is below the teddy bear} & \supl{an orange teddy bear and a purple skateboard, the teddy bear is below the skateboard, in a sunlit room with bare walls} \\[14pt]
\rotatebox{90}{\tiny\ Gemini: right} & \supimg{\includegraphics[width=\supw]{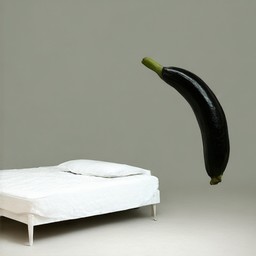}}{oiGreen!85!black}{\checkmark} & \supimg{\includegraphics[width=\supw]{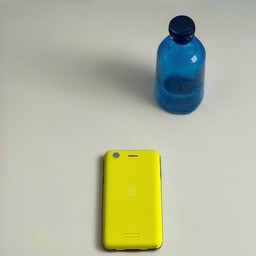}}{oiGreen!85!black}{\checkmark} & \supimg{\includegraphics[width=\supw]{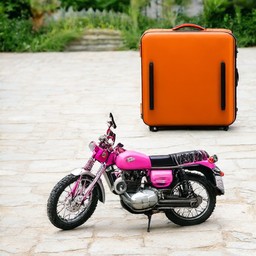}}{oiGreen!85!black}{\checkmark} & \supimg{\includegraphics[width=\supw]{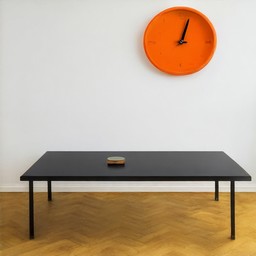}}{oiGreen!85!black}{\checkmark} & \supimg{\includegraphics[width=\supw]{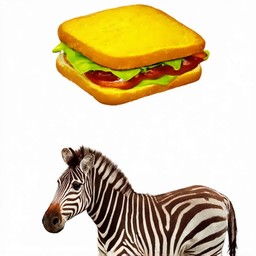}}{oiGreen!85!black}{\checkmark} & \supimg{\includegraphics[width=\supw]{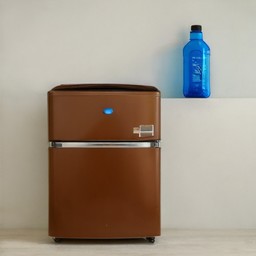}}{oiGreen!85!black}{\checkmark} \\[1pt]
 & \supl{a white bed and a black banana, the bed is below the banana} & \supl{a yellow cell phone and a blue bottle, the cell phone is below the bottle} & \supl{a pink motorcycle and an orange suitcase, the motorcycle is below the suitcase, on a stone plaza} & \supl{a black dining table and an orange clock, the dining table is below the clock, on a wooden floor} & \supl{a brown zebra and a yellow sandwich, the zebra is below the sandwich} & \supl{a brown refrigerator and a blue bottle, the refrigerator is below the bottle} \\[14pt]
\end{tabular}
\caption{\textbf{Gemini audits of checker-accepted teacher images.} (a) The box-accepted ``on top of'' targets of the $2\times2$ (Sec.~5.3 of the main paper): the first six prompts by ID that Gemini calls floating, after the three of Fig.~4 of the main paper, and the first six it does not; badges give Gemini's full verdict under the split verifier's rule (on top, touching, not floating, not fused). (b) The checker-passed compound ``below'' images of \cref{app:below}: the first six by ID that Gemini rejects and the first six it fully accepts (three images with unparsed Gemini answers are left out). The rules were fixed before any image was viewed; badges give Gemini's verdicts, which no human checked.}
\label{fig:audit}
\end{figure*}

In a small post-hoc check, the teacher drew 361 compound prompts (a position prompt with two colored objects,
or a count prompt with an added colored object), and Gemini reviewed the 273 images the checker passed.
Of the 57 ``below'' images, Gemini judged 13 (22.8\%) wrong; for 3 more its answers were incomplete.
The typical case: the teacher places the two colored objects side by side
on the ground, and one box center sits slightly lower, so the checker reads ``below''. The sample is
small, and we treat it only as supporting evidence for Sec.~5.3 of the main paper. %
\cref{fig:audit} shows images from this check and from the Gemini audit of the $2\times2$ targets.

\end{document}